\documentclass{article}

\usepackage{PRIMEarxiv}

\usepackage[utf8]{inputenc} 
\usepackage[T1]{fontenc}    
\usepackage{hyperref}       
\usepackage{url}            
\usepackage{booktabs}       
\usepackage{amsfonts}       
\usepackage{nicefrac}       
\usepackage{microtype}      
\usepackage{lipsum}
\usepackage{colortbl}
\usepackage{pgfplots}
\usepackage{amsmath}
\usepackage{pgfplotstable}
\usepackage{filecontents}
\pgfplotsset{compat=1.18}
\pgfplotsset{width=10cm,compat=1.9}
\usepackage{amsmath}
\usepackage{multirow}
\usepackage{xcolor}
\usepackage{algorithm}
\usepackage{pgfplots}
\pgfplotsset{compat=1.18}
\usepackage{pgfplotstable}
\usepackage{filecontents}
\usepackage{algpseudocode}
\usepackage{booktabs}
\usepackage{fancyhdr}       
\usepackage{graphicx} 
\usepackage{subcaption}
\graphicspath{{media/}}     
\newcommand{\bs}[1]{\boldsymbol{#1}}
\newcommand{\ceil}[1]{\left\lceil #1 \right\rceil}

\definecolor{blue1}{RGB}{0, 82, 147}     
\definecolor{blue2}{RGB}{47, 128, 181}   
\definecolor{blue3}{RGB}{158, 202, 225}  

\definecolor{red1}{RGB}{192, 57, 43}     
\definecolor{red2}{RGB}{214, 108, 89}    
\definecolor{red3}{RGB}{232, 175, 163}   
\title{On the under-reaching phenomenon\\ in message passing neural PDE solvers: \\revisiting the CFL condition}

\author{
  L. Tes\'an\textsuperscript{1,*}, M. M. Iparraguirre\textsuperscript{1,*},  D. Gonz\'alez\textsuperscript{1}, P. Martins\textsuperscript{1,2}, E. Cueto\textsuperscript{1}\vspace{0.3cm}  \\ 
  \textsuperscript{1} ESI Group-UZ Chair of the National Strategy on Artificial Intelligence. \\ Aragon Institute of Engineering Research (I3A),\\ Universidad de Zaragoza. Zaragoza, Spain. \\
  \textsuperscript{2} Aragonese Foundation for Research and Development (ARAID), Aragón, Spain.\\ \vspace{0.3cm}
  \textsuperscript{*} Equal contributors.\vspace{0.2cm}
  \\ 
  \texttt{\{ltesan,mikel.martinez,gonzal,pmartins,ecueto\}@unizar.es}
}

\begin{document}
\maketitle

\begin{abstract}

This paper proposes sharp lower bounds for the number of message passing iterations required in graph neural networks (GNNs) when solving partial differential equations (PDE). This significantly reduces the need for exhaustive hyperparameter tuning. Bounds are derived for the three fundamental classes of PDEs (hyperbolic, parabolic and elliptic) by relating the physical characteristics of the problem in question to the message-passing requirement of GNNs. In particular, we investigate the relationship between the physical constants of the equations governing the problem, the spatial and temporal discretisation and the message passing mechanisms in GNNs.

When the number of message passing iterations is below these proposed limits, information does not propagate efficiently through the network, resulting in poor solutions, even for deep GNN architectures. In contrast, when the suggested lower bound is satisfied, the GNN parameterisation allows the model to accurately capture the underlying phenomenology, resulting in solvers of adequate accuracy. 

Examples are provided for four different examples of equations that show the sharpness of the proposed lower bounds.

\end{abstract}

\keywords{Neural PDE solvers, message passing, graph neural networks, partial differential equations.}

\section{Introduction}

Machine learning-based simulators have gained significant attention due to their ability to approximate complex physical phenomena at a fraction of the computational cost of traditional numerical solvers \cite{montans2023machine,pfaff2021learning,aldakheel2025physics}. Once trained, these learned simulators can generate rapid predictions, making them particularly useful for tasks requiring real-time inference \cite{FlowGNN}, inverse problem-solving \cite{InverseGal} or chaotic-spatial dependent systems \cite{lam2022graphcast}.   

Traditional physics-based solvers rely on mechanistic models, typically expressed as partial or ordinary differential equations (PDE or ODE, respectively) and discretized into algebraic problems suitable for computation. Their accuracy hinges on the fidelity of the physical model and the discretization used as solver, such as finite differences \cite{jordan1965calculus}, finite elements \cite{bathe2007finite}, or finite volumes \cite{eymard2000finite}. These traditional approaches often demand substantial computational resources due to their limited ability to capture complex phenomena efficiently. To achieve accurate solutions, they require very fine spatial discretization and high temporal resolution, which significantly increases the number of iterations, and consequently their computational cost.

Neural Networks (NNs) have emerged as powerful simulation tools, enabling the development of solvers through two main paradigms. These are: physics-informed learning, where the residuals of the governing PDEs are minimized directly, possibly without the need for training data \cite{Karniadakis2021}; and data-driven learning, where models are trained on simulation or experimental data and subsequently used to predict solutions without explicitly solving the underlying equations at inference time \cite{Zhao2024}. In the data-driven approach, performance relies heavily on the quality and diversity of the training data, as well as the expressive capacity of the chosen neural architecture \cite{pfaff2021learning, allen2022physicaldesignusingdifferentiable, allen2022learningrigiddynamicsface}. While training can be computationally demanding, inference is typically very efficient \cite{CuetoMikel}.

Physics-Informed Neural Networks (PINNs) emerge as the primary alternative for scenarios where the governing equations are known \cite{Karniadakis2021, Cuomo2022, Cai2021}. This approach solves partial differential equations by embedding physical laws into the loss function, enabling solution approximation without labeled data or discretization. They are mesh-free and can directly optimize PDEs thanks to the differentiability of neural networks. However, they share their limited flexibility with traditional solvers—any modification to initial or boundary conditions requires re-optimizing the problem from scratch, and they generally lack the ability to generalize to unseen problems \cite{Kim_Lee_Lee_Jhin_Park_2021, fesser2023understandingmitigatingextrapolationfailures}.
Despite the training challenges of this approach, PINNs can minimize the PDE at a limited number of collocation points, while enabling high-resolution simulations during inference by evaluating the model over a dense set of points. This represents a significant advantage over traditional solvers.

On the other hand, models based on dimensionality reduction techniques can condense the complex hidden dynamics within data into lower-dimensional representations. This facilitates the training of neural simulators and establishes a hybrid synergy between human expertise and artificial intelligence. However, they still suffer from the same fundamental limitations: an inability to locally interpret the underlying physics of the problem, and consequently, a lack of extrapolation capability to new geometries or domains \cite{Fresca2021, fresca-PODL}. Nevertheless, promising new research directions are emerging that enable the incorporation of domain-specific information, offering equally interesting alternatives for overcoming these challenges \cite{Fresca2025}.

GNNs have emerged in the past years as one of the most powerful data-driven approaches for modeling physical phenomena. Similar to how convolutional networks \cite{lecun1998gradient} perform local optimization by acting as spatial operators capable of learning complex patterns \cite{Bronstein-2021_geodistnce, bronstein2017geometric}, GNNs leverage their architecture to capture both local and global interactions within graph-based representations, such as mesh discretizations used in physical simulations. By establishing mathematical relationships between spatial discretization and the underlying physical processes, they can effectively represent nearly any type of phenomenology, regardless of its
complexity \cite{GNNsurvey}. In pursuit of explainability and theoretical guarantees, an important extension of these models is Physics-Informed GNNs (PIGNNs), which enhance their capabilities by embedding physical laws such as conservation principles, symmetries, or governing partial differential equations directly into their architectures or loss functions \cite{InverseGal, NEURIPS2022_17b598fd, LIU2023211486}. This integration reflects the broader paradigm of hybrid artificial intelligence, where data-driven models are guided by domain knowledge. Alternatively, other hybrid approaches adopt a thermodynamic perspective, incorporating constraints inspired by thermodynamic principles \cite{TIGNNs}. By framing the phenomenon in a thermodynamic setting, these models offer a flexible framework for modeling open, non-equilibrium, or multiscale systems, while increasing interpretability \cite{hernandez2021structure,hernandez2023port,hernandez2023thermodynamics,tierz2025graph,gruber2023reversible,moya2023thermodynamics,bermejo2024thermodynamics,tierz2025graph}.

Among the many factors that influence the success of deep learning models, hyperparameter tuning plays a particularly critical role. It directly affects the model's ability to learn effectively and generalize well. Poor choices in hyperparameter settings can lead to suboptimal training outcomes and degraded performance. This tuning process---an essential component of model selection---typically demands significant computational resources and time. Common hyperparameters include learning rate, number of hidden units, and batch size, which are often selected through trial-and-error strategies. Guided search methods \cite{FeurerHutter} can help identify optimal configurations more efficiently. Nevertheless, hyperparameter tuning remains a challenge. In GNNs, a particularly critical hyperparameter is the number of message-passing iterations (MPI), which directly affects the model's ability to propagate information across the graph. While this has traditionally been treated as a standard tunable parameter, we propose establishing a lower bound based on the underlying system dynamics to reduce the hyperparameter search space and ensure sufficient information flow for accurate prediction \cite{pmlr-v139-balcilar21a}.

To estimate the lower bound on MPI, we distinguish between two regimes. Hyperbolic systems (e.g., transport or wave equations~\cite{lax1973hyperbolic}), are characterised by the presence of a wave which, in one form or another, propagates through the domain, so the proposed lower bound reflects the speed at which this wave travels. In parabolic and elliptic systems, in contrast, the described phenomena propagate with infinite speed, affecting the whole domain instantaneously---making the lower bound a function of the system's geometry and mesh size. In particular, for systems governed by elliptic equations, it is well known that they predict, contrary to experience, an infinite transmission rate. This phenomenon is called the infinite propagation speed paradox \cite{auriault2017paradox}. These principles apply to both standard message-passing networks~\cite{Gilmer2020} and more advanced models like MeshGraphNets~\cite{pfaff2021learning}. 

By aligning hyperparameter selection with the underlying physics of the system, our approach ensures that learned graph simulations adhere to the principles of numerical stability while maintaining computational efficiency. This bridges a critical gap between data-driven modeling and the rigorous demands of scientific computing.

The main contribution of this article, therefore, is the proposition of a lower bound for the number of message steps as a function of the character of the equation governing the physical phenomenon. As will be seen below, the proposed bound for hyperbolic equations is strongly reminiscent of the famous Courant-Friedrichs-Lewy condition for finite differences applied to this type of equations. In problems governed by elliptic and parabolic equations, on the contrary, the proposed bound must ensure that the information about the phenomena taking place reaches all corners of the domain. It will be seen, therefore, how the bounds are very different in one case and the other, but how the predictions show great accuracy if they are fulfilled.

The layout of the paper is as follows. In Section \ref{GNNs} we review the basics of message-passing graph neural networks for the solution of PDEs. In Section \ref{state} we review the problems arising from the insufficient number of message steps, a phenomenon known as under-reaching in the literature \cite{lu2024nodemixup} and develop lower bounds to ensure that this problem does not occur. In Section \ref{numres} numerical tests are described which show the validity of the proposed limits. We follow in Section \ref{datasets} with the description of the examples used for analysis of the dynamics of message passing, while in Section \ref{experiments} we show the results obtained. The paper is closed with Section \ref{conclusions} devoted to deriving the conclusions of this work.

\section{Message-passing graph neural networks}\label{GNNs}


Our model architecture, with which we have performed the experiments to be shown in the following sections, consists of a one-step predictor, except for elliptic problems, where the time variable is not present. The  prediction at each time step becomes the current state used as input for the next time step. This setup makes the model easy to train, as it only requires training on one-step predictions, although more complex architectures exist, involving multi-step temporal integration, auto-regressive schemes, etc. \cite{brandstetter2022message}. Additionally, it is flexible during inference, allowing us to simulate long time periods ($n$-step rollouts) without explicitly requiring time as a parameter.

\begin{figure}[h!]
\centering
\includegraphics[width=\textwidth,trim=0 600 0 100, clip]{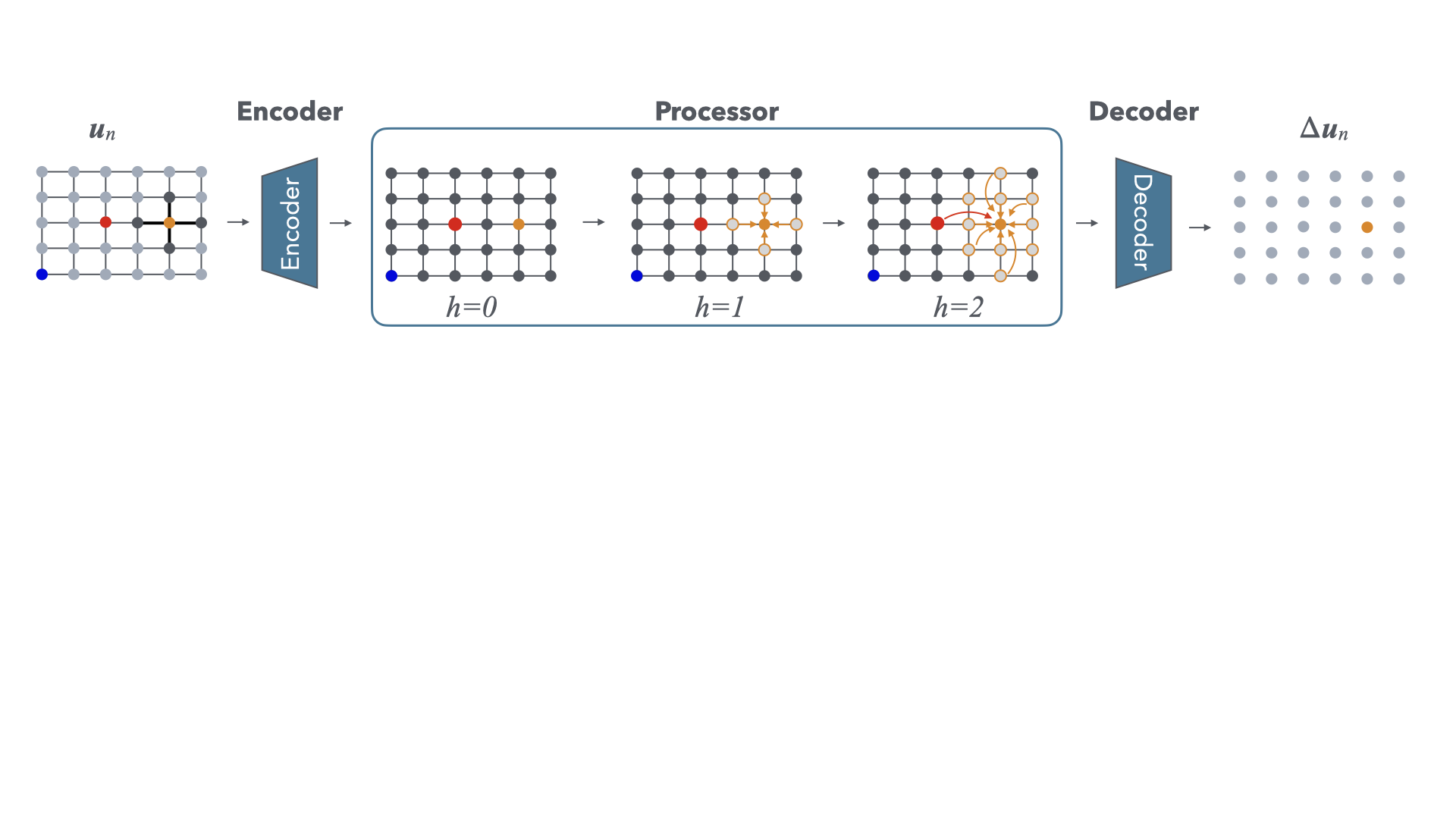}
\caption{Illustration of a GNN with an encoder-processor-decoder approach and $h=1, \ldots, M$ passes---we depict the case $M=2$---. We focus on the orange node of the graph, whose evolution from time step $n$ to $n+1$ is of primary interest. This evolution is driven by the perturbations originating from the red (located at a distance of two hops, $h=2$) and blue nodes (located at a distance of six hops, $h=6$). Therefore, it is crucial for the processor, through message passing iterations, to propagate the information from the red and blue nodes to the orange node. If the number of iterations is insufficient (less than the $h$-distance between nodes), the orange node's future state will not be estimated based on the actual nodes that influence it, but rather inferred from an incomplete or indirect representation. This can lead to suboptimal or incorrect predictions.}
\label{fig:gnn_reach}
\end{figure}

This one-step predictor is a GNN that follows the encoder-processor-decoder architecture, which has demonstrated significant potential in modeling mesh-based and physical problems, see Fig. \ref{fig:gnn_reach}. For the theoretical developments and the first numerical examples we wanted to keep the network architecture as simple as possible, in order to eliminate the possible influence of other factors in the analysis. In Section \ref{sec:lower-bound-geom}, however, a more complex architecture is used, based on MeshGraphNets \cite{pfaff2021learning}, the details of which will be explained in the section itself. For the time being, let us stick to the simplest possible architecture, which does not include any encoding on the edges of the graph and only stores and processes nodal variables:
\begin{itemize}
    \item Every node in the graph shares a multi-layer perceptron that encodes the input \( \bs{u}_{{n}} = \bs u (t = n\Delta t) \in  \mathbb R^{\tt n_{\text{dof}}} \) into a high-dimensional latent space  \( \bs{\xi}_{{n}}  \in  \mathbb R^D \), with $D>{\tt n_{\text{dof}}}$, in general. Here, ${\tt n_{\text{dof}}}$ represents the number of state variables at each node.
    \item The processor, leveraging a message-passing algorithm, aggregates information from neighboring nodes within a hop distance \( h \) at each iteration. In this stage, the GNN updates the graph \( G \) in the latent space.  
    \item The decoder transforms the nodal representations from the latent space \(\bs{\xi}_{{n+1}} \) back into the physical space, as the prediction \( \Delta \bs{u}_{{n}} \) per node.  
\end{itemize}

The GNN is trained to predict one-step system evolutions, \( \Delta \bs{u}_n \), which depend solely on the input \( \bs{u}_n \) and the learned network parameters \( \boldsymbol{\theta} \),
\begin{equation}
    \bs{u}_{n+1} = \bs{u}_{n} + \Delta \bs{u}_{n}, 
\label{eq:integrator}
\end{equation}
where $ \Delta \bs{u}_{n} =  \text{GNN}(\bs{u}_{n}, \boldsymbol{\theta})$.

The processor is the core of this type of neural solver. In essence, at each time step $n$---which we omit for clarity---, it performs the learning of $h=1, \ldots, M$ message passing processes of the form
\begin{itemize}
\item Learn the message along edge $ij$: 
\begin{equation}\label{mssg1}
\bs m_{ij}^h = \phi (\bs \xi_i^h, \bs \xi_j^h,\bs \theta),
\end{equation}
\item Update of node $i$: 
\begin{equation}\label{mssg2}
\bs \xi_i^{h+1} = \varphi \left(\bs \xi_i^h, \sum_{j\in \mathcal N(i)}\bs m_{ij}^h, \bs \theta \right),
\end{equation}
\end{itemize}
where $\mathcal N(i)$ represents the neighborhood of node $i$, i.e., the set of nodes $j$ connected by the graph to node $i$.

The ability of GNNs to understand and process global information from the graph \( G \), despite being a nodal level architecture, is entirely dependent on the transmission of the message passing algorithm. If the number of messages $M$ is insufficient, 
the nodes are in practice disconnected, which prevents effective information propagation. This limitation can compromise the model's capacity to capture regional or global dependencies necessary for accurately solving our problem. Even in a connected graph, inadequate message passing prevents effective communication between nodes, leading to suboptimal performance.

Previous work has shown that solvers based on message passing mechanisms have the ability to reproduce classical PDE solving schemes such as finite difference, finite volume or WENO-type schemes (schemes with $M=1$, $M=2$ and $M=3$, respectively), see \cite{brandstetter2022message}. However, the authors of the mentioned paper use a higher number of passes, six hops. Additionally, at each hop they consider a larger number of neighbors for each node than would be indicated by the connectivity of the graph, using up to twenty neighbors per node. The result, if translated into a finite difference scheme, would result in a much higher order scheme and a highly populated stiffness matrix.

In general, almost all existing literature practices ``many rounds of message passing'' \cite{lam2022graphcast}. Pfaff and coworkers state that ``increasing the number of graph net blocks (message passing steps) generally improves performance, but it incurs a higher computational cost'', show that increasing the number of passes implies a decreasing error in the prediction and generally employ fifteen hops as a compromise between computational cost and accuracy \cite{pfaff2021learning}. In general, this phenomenon is now coined as under-reaching and refers to a situation where a node cannot detect or recognize nodes located beyond its specified number of layers \cite{lu2024nodemixup}.

If this were literally the case, the number of passes necessary to achieve the maximum accuracy of the method would be the number of passes necessary for the messages to get from one end of the mesh to the other (we do not consider the possibility of using extended neighborhoods as in \cite{brandstetter2022message}, to alleviate this process, in order to simplify the hypotheses as much as possible). As will be seen in the following sections, this is not always the case. It is not always necessary for messages to reach from one end of the graph to the other, and in any case, this depends on the type of PDE that governs the physical phenomenon under study.

\section{The phenomenon of under-reaching seen from a physical point of view}\label{sec:problem_statement}\label{state}

In this section, we investigate the impact of the number of message-passing iterations in GNNs across the three categories of partial differential equations: hyperbolic (Section \ref{sec:hyperbolic_theory}), parabolic (Section \ref{sec:parabolic_theory}), and elliptic (Section \ref{sec:elliptic_theory}). Understanding the distinct characteristics of the physical phenomena described by each PDE family is crucial, as it reveals how information should be propagated across the graph. To this end, we analyze representative benchmark problems for each class: the wave equation (hyperbolic), the Fourier heat transfer equation (parabolic), and the Poisson equation for electrostatic potential (elliptic). Additionally, we include a more complex, real-world scenario involving incremental forming:  a metal sheet is deformed by a tool, assumed to be rigid, into the required shape.

\subsection{Phenomena governed by hyperbolic equations}\label{sec:hyperbolic_theory}

A defining characteristic of hyperbolic problems is the presence of a wavefront---a localized disturbance that propagates through the domain over time---. This reflects the inherently local and directional nature of information flow in such systems, which poses specific challenges and requirements for message-passing in GNNs. One of the most well-known examples of a hyperbolic PDE is the unimodal wave equation.

The wave equation models phenomena such as sound waves, electromagnetic waves, and vibrations in a string. The general form of the wave equation in two spatial dimensions (2D) can be written as:
\begin{equation}
\frac{\partial^2 u}{\partial t^2} = c^2 \left( \frac{\partial^2 u}{\partial x^2} + \frac{\partial^2 u}{\partial y^2} \right) =  c^2 \nabla^2 u, \quad \text{in } \Omega \times (0,T],
\label{eq:wave_equation}
\end{equation}
where \( u(x, y, t) \) represents the wave's height in a two-dimensional domain, and the wave propagates through both the \( x \)- and \( y \)-directions with speed \( c \). In this scenario, the physical term that governs our problem is the wave propagation speed \( c \), a constant that will be essential in defining the lower bound of the MPIs.

Since, as we have seen, some authors have demonstrated the ability of GNNs using message passing to learn finite-difference type schemes from data, it is useful to study these message passing processes in the light of the physical constraints that apply on finite-difference schemes, particularly, the Courant-Friedrichs-Lewy condition \cite{de2013courant}.

To numerically approximate the solution of the described hyperbolic problem, we discretize the computational domain into a uniform spatial grid with spacing $\Delta x$ and $\Delta y$, respectively, while time is discretized with a time step $\Delta t$. 
For the 2D finite difference scheme to be stable, the time step \( \Delta t \) the famous Courant-Friedrichs-Lewy (CFL) condition establishes that, for a given \( c_x \), \( c_y \), \( \Delta x \), and \( \Delta y \) mesh resolution in space:
\[
\frac{c_x \Delta t}{\Delta x} + \frac{c_y \Delta t}{\Delta y} \leq 1, 
\]
while for a square grid mesh (\( \Delta y = \Delta x \)) with constant radial speed, 
$$ \frac{c \Delta t}{\Delta x} \leq \frac{1}{\sqrt{2}}.
$$
What the CFL condition is imposing on the discretization scheme is, plain and simple, that the wave described by Eq.~(\ref{eq:wave_equation}) does not traverse more than the distance between two nodes in a single time increment. Otherwise, the scheme will be unstable and its accuracy will decline rapidly.

In other words, what the CFL condition is imposing is that, in each time increment, the physical wave does not go beyond the immediate neighborhood of each node. If we transfer this way of reasoning to the message passing scheme, what we would have is that the wave cannot reach, in each time increment, a greater distance than the message reaches. In other words, assuming for simplicity and without loss of generality a squared mesh, 
\begin{equation*}\label{CFL2}
M\Delta x > \sqrt{2} c\Delta t,
\end{equation*}
or
\begin{equation}\label{CFL3}
M > \sqrt{2}\frac{c\Delta t}{\Delta x}.
\end{equation}
In general, this figure should be complied with, suitably rounded to a whole number of passes, so that
\begin{equation}\label{CFL4}
M > \ceil{\sqrt{2}\frac{c\Delta t}{\Delta x}},
\end{equation}
where $\ceil{\cdot}$ indicates the ceiling function.

\subsection{Phenomena governed by parabolic equations}\label{sec:parabolic_theory}

A parabolic PDE models diffusion processes, such as heat conduction and chemical diffusion, where the state of the system evolves over time. This group of problems is characterized by a diffusivity term, which causes perturbations to spread across the entire domain. The evolution of the system occurs at a rate determined by the flux field, which changes as the system approaches equilibrium. In contrast to hyperbolic problems, where a wavefront propagates the perturbation, parabolic problems involve a simultaneous, diffusive perturbation that affects the entire domain at each time step. This is the well-known paradox of infinite propagation velocity in diffusing systems \cite{bertola2007speed}.

The ubiquitous example of a parabolic problem is the heat transfer equation, which describes the distribution of temperature over time in a given domain. Consider the Fourier heat equation on a unit square domain $\Omega$ as a model for diffusive parabolic PDEs:
\begin{equation}
\frac{\partial u}{\partial t} = \alpha \nabla^2 u \quad \text{in } \Omega \times (0,T],
\end{equation}
where $u(x,y,t)$ represents now the temperature distribution, $\alpha > 0$ is the thermal diffusivity constant, and $\nabla^2 = \frac{\partial^2}{\partial x^2} + \frac{\partial^2}{\partial y^2}$ is the Laplacian operator. The problem is completed with appropriate initial and boundary conditions.

When numerically approximating  parabolic partial differential equations, given the paradox of infinite propagation velocity, the entire domain must be informed about the origin of the perturbation at each temporal step \cite{auriault2017paradox}. This means that the state at every node in the domain is influenced by all other nodes. In other words, each point must ``see'' the global state of the system to correctly capture the underlying physical behavior. This observation motivates our design of the lower bound for message passing iterations in such systems. To ensure that the influence of boundary conditions and global geometric features is properly captured, the number of message passing iterations must be sufficient for information to propagate throughout the domain,
\begin{equation}\label{eq:gometric_bound}
M =  \frac{L}{\Delta x},
\end{equation}
where \( L \) denotes the size of the domain, \( \Delta x \) is the spatial discretization, and \( M \) represents the number of hops required to propagate information from one boundary of the domain to the opposite side. In this work, we assume, for simplicity, that all domains are square, with uniform spatial discretization such that \( \Delta x = \Delta y \). This hypothesis, which might at first sight seem excessively rigid, is motivated by our desire not to mix another problem of graph networks, the so-called over-squashing \cite{toppingunderstanding}. The number of $h$-hop neighbors grows exponentially with $h$. If, in addition and due to the topology of the network, bottlenecks occur, the transmission of messages can become computationally very expensive. In order to isolate this effect conveniently, as mentioned above, we have preferred to analyze examples that are developed in square domains and that, therefore, do not present these difficulties.

Our principal hypothesis states that the optimal hyperparameter configuration for the GNN must guarantee that the information carried by the messages reaches the farthest side of the domain. This condition ensures that each inference step incorporates a global understanding of the system's behavior.

\subsection{Phenomena governed by elliptic equations}\label{sec:elliptic_theory}

Elliptic phenomena are typically associated with steady-state processes, where the system has reached equilibrium, and there is no time dependence. These problems are characterized by partial differential equations that describe spatial relationships between variables, such as elasticity or potential in a given domain. Unlike hyperbolic and parabolic problems, elliptic equations do not involve wave propagation or diffusion over time but instead focus on spatial distributions that satisfy a given condition across the entire domain.

As in the elliptic case, problems governed by parabolic equations require messages to reach the far ends of the grid, which, in general, will add significant computational cost to the problem. Therefore, the stability condition will revert to that expressed by Eq. (\ref{eq:gometric_bound}).

As an example, we consider, on one side, the electrostatic potential $u(x,y)$ for a two-dimensional domain $\Omega \subset \mathbb{R}^2$, governed by {Poisson's equation} in the presence of a charge distribution $\rho(x,y)$:
\begin{equation}
\nabla^2 u = -\frac{\rho}{\epsilon_0},
\label{eq:poisson}
\end{equation}
where $\epsilon_0$ is the permittivity of free space,  $\rho(x, y)$ is the charge density distribution and $\nabla^2 = \frac{\partial^2}{\partial x^2} + \frac{\partial^2}{\partial y^2}$ is the Laplacian operator. Again, the problem is completed with appropriate boundary conditions.

\begin{figure}[htbp]
    \centering        
    \includegraphics[width=1.0\linewidth]{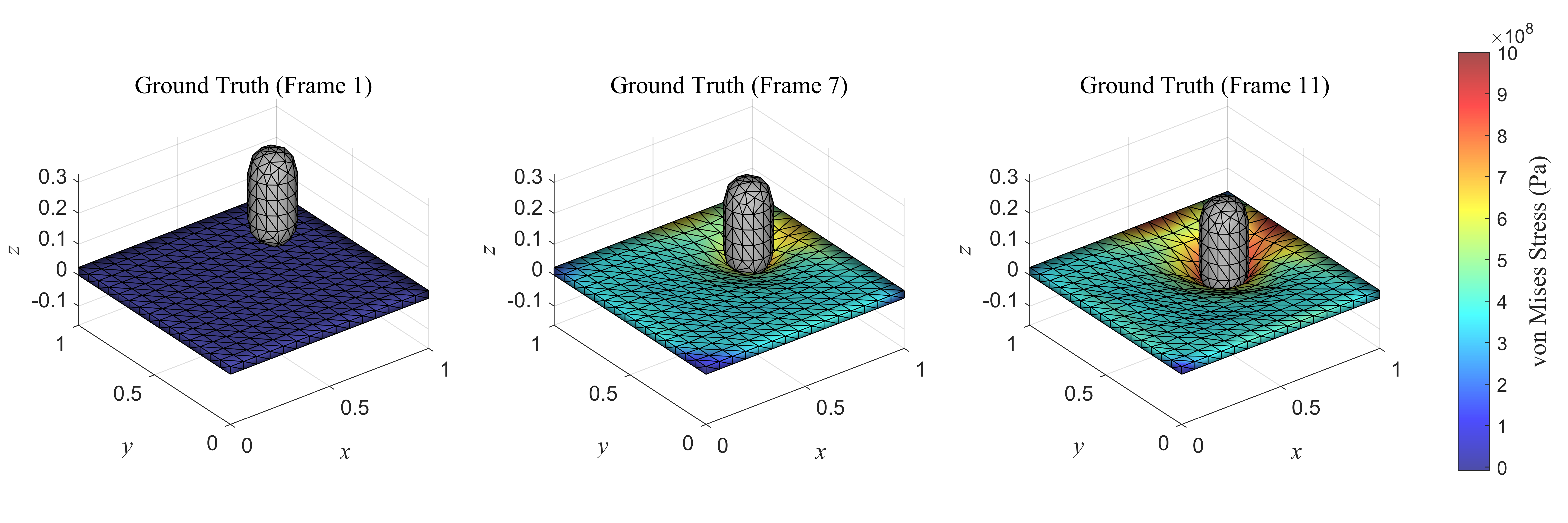}
    \caption{Evolution of the von Mises stress for the  incremental forming problem simulated with finite element methods. The FEM simulations used 1000 time steps, of which only 10 are saved for training. The actuator impacts the plate from both upper and lower \(z\)-directions, producing plastic deformation. Clamped boundary conditions are applied at plate edges. Time steps 1, 7 and 11 of one of the data sets are shown, respectively, from left to right.}
    \label{fig:example-ellip-plate}
\end{figure}

As a second elliptic example, we model quasi-static incremental forming process, that involves the collision between a rigid tool and a elasto-plastic square plate in 3D space, see Fig. \ref{fig:example-ellip-plate}. This setup introduces additional complexity, including the 3D nature of the problem, the use of multivariate fields---namely the displacement vector \( \boldsymbol{u} \) and stress tensor \( \boldsymbol{\sigma} \)---and the presence of contact interactions between two bodies. In this context, the resulting deformations depend on the boundary conditions, the geometry of the system, and the magnitude and location of the imposed displacements from the actuator.

The governing partial differential equation for elasto-plastic phenomena in 3D is the equilibrium equation, for which we assume large strain settings,
$$
\bs \nabla_0 \bs P + \bs f_0 = \bs 0,
$$
where $\bs P$ represents the first Piola-Kirchhoff stress tensor, and $\bs \nabla_0 = \frac{\partial \bs P}{\partial \bs X}$, with $\bs X$ the reference configuration of the solid. $\bs f_0$ represents the body forces at the reference configuration. This equation must be supplemented with appropriate Dirichlet and Neumann boundary conditions. In addition, we assume that the material follows the classical multiplicative decomposition of the deformation gradient into elastic and plastic parts,
$$
\bs F = \bs F^{el}\cdot \bs F^{pl},
$$ 
with rate-independent plasticity, beginning when the yield function 
$$
f(\bs \sigma, H) <0,
$$
---considered here as J2 plasticity---is reached. $H$ represents the (isotropic) hardening parameter of the constitutive model and $\bs \sigma$ the Cauchy stress tensor.

Since any of the above phenomena-and, in general, any other governed by elliptic equations-are essentially static, time plays no role in the problem and therefore information must be transmitted instantaneously throughout the domain. This results in a lower bound on the number of passes which is given, again, by Eq. (\ref{eq:gometric_bound}).

\section{Numerical results}\label{numres}

\subsection{Synthetic datasets}\label{datasets}

We have generated four distinct sets of data, one for each of the problems described in Section \ref{sec:problem_statement}. Each dataset consists of 1,000 simulations, which are split into 80\%, 10\%, and 10\% for training, validation, and testing, respectively. 

As mentioned before, the hyperbolic and parabolic cases are generated with finite differences, while the elliptic case is solved with finite element methods (FEM). Both of these methods require fine spatial and temporal discretizations to ensure stability and convergence, leading to high computational costs due to the large number of iterations that the solvers must comply with. To address this cost, we pruned the spatial and time domains at a coarser ratio, where these methods would typically fail. As will be shown, message-passing GNNs show their potential in solving complex physical problems with coarser spatial and temporal discretizations.

To rigorously assess the impact of the number of message-passing iterations, we generate two datasets for each problem, intentionally varying the number of iterations required for convergence. This variation is achieved by altering the spatial resolution, thereby producing datasets of different granularities, as summarized in Table~\ref{tab:tab_datasets}.

\begin{table}[h!]
\centering
\begin{tabular}{|l|c|c|c|c|c|c|c|c|c|c|}
\hline
\textbf{Dataset} & ${\Delta x}$ & ${\Delta t}$ & $M$ & $\times N$ & $c$ & \text{Nodes} & \text{Discretiz.} & \text{Dim.} & \text{Solver} & $L\times W (\times H) $ \\ \hline \hline

\multicolumn{11}{|l|}{\textbf{Hyperbolic}, Fig. \ref{fig:example-2datasets}} \\ \hline 
$\text{Wave-LowRes }$ & 0.04 & 0.001 & 4 & 200 & 0.5 & 625 & Rect. & 2D & FD & 1 $\times$ 1 \\ 
$\text{Wave-HighRes}$ & 0.02 & 0.001 & 8 & 200 & 0.5 & 2500 & Rect. & 2D & FD & 1 $\times$ 1 \\  \hline\hline

\multicolumn{11}{|l|}{\textbf{Parabolic}} \\ \hline
$\text{Fourier-LowRes}$ & 0.1 & 0.0004 & 10 & 200 & -- & 100 & Rect. & 2D & FD & 1 $\times$ 1 \\ 
$\text{Fourier-LowRes}$ & 0.1 & 0.0004 & 20 & 200 & -- & 400 & Rect. & 2D & FD & 2 $\times$ 2 \\ \hline\hline

\multicolumn{11}{|l|}{\textbf{Elliptic}, Fig. \ref{fig:example-ellip-plate}} \\ \hline
$\text{Poisson-LowRes}$ & 0.1 & -- & 10 & 500 & -- & 100 & Rect. & 3D & Jacobi & 1 $\times$ 1 \\
$\text{Poisson-HighRes}$ & 0.05 & -- & 20 & 500 & -- & 400 & Rect. & 3D & Jacobi & 1 $\times$ 1 \\
$\text{Plastic Plate}$ & 0.065 & 0.001 & 15 & 100 & -- & 625 & Tethr. & 3D & FEM & 1 $\times$ 1 $\times$ 0.1 \\ \hline
\end{tabular}
\vspace{0.2cm}
\caption{Dataset characteristics and numerical methods. Hyperbolic and parabolic cases use finite differences (FD) on 2D rectangular meshes; elliptic cases use finite element methods (FEM) on 3D tetrahedral meshes. Columns show: dataset name, spatial step \(\Delta x\), time step \(\Delta t\), estimated lower-bound message passing iterations $M$, time sampling $N$, wave speed \(c\) (hyperbolic only), mesh nodes, discretization type, problem dimension, numerical solver, and domain dimensions. For the plastic plate problem, $\Delta t$ represents the pseudo-time through which the loading process is discretized. The problem is actually quasi-static and the load is applied within a unit time period.
}
\label{tab:tab_datasets}
\end{table}

\begin{figure}[htbp]
    \centering
    \includegraphics[width=1.0\linewidth]{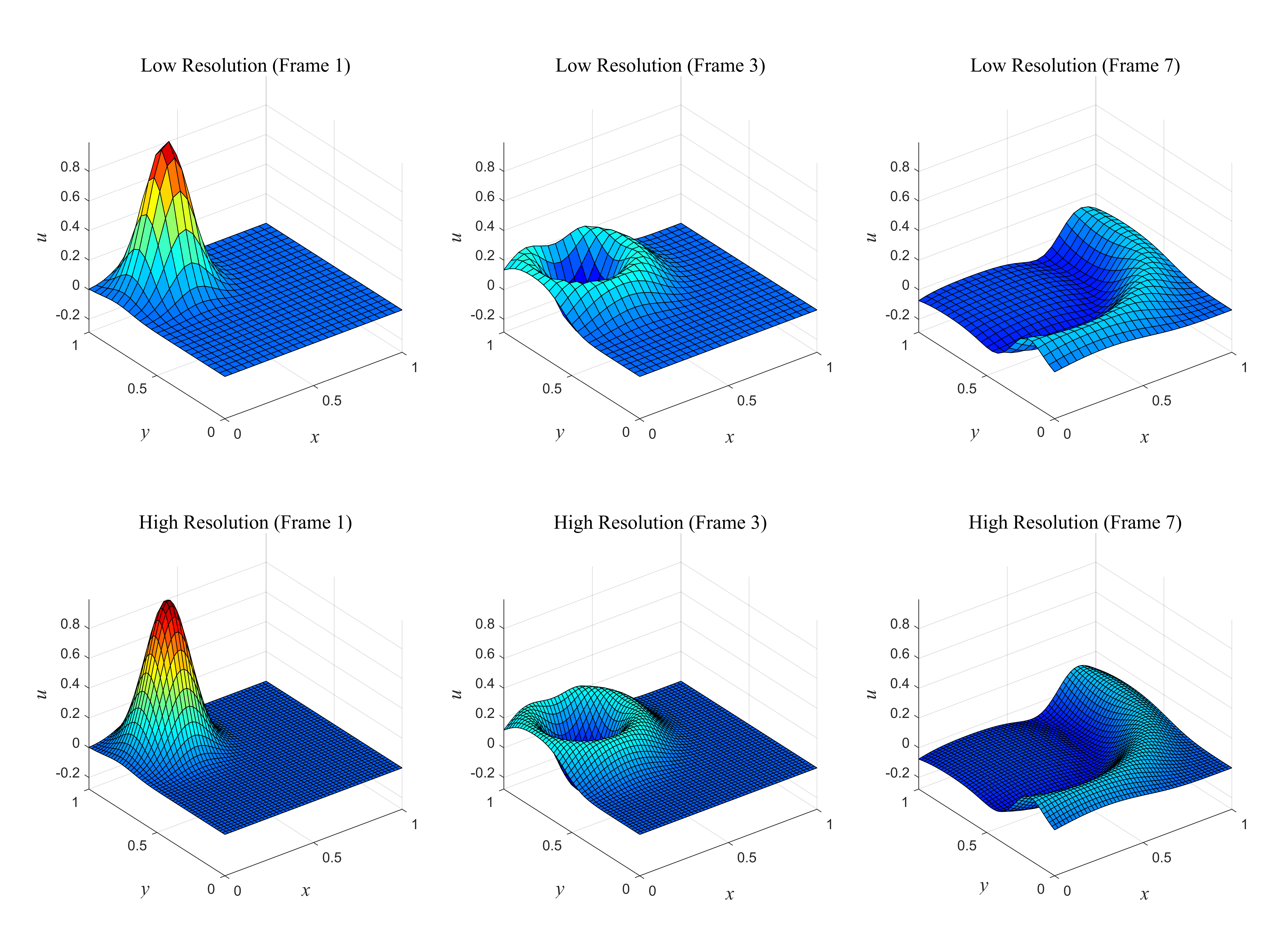}
    \caption{Temporal evolution of one of the wave problems computed using finite difference methods with two distinct spatial resolutions (low and high---top and bottom rows, respectively). Each simulation uses a temporal sampling of \( N = 200 \) time steps, corresponding to a total duration of \( 2.0 \) seconds. These time steps are pruned and only ten of them kept for training. The wave speed is fixed at $c=0.5$. Time steps 1, 3 and 5 are represented at the left, center and right columns, respectively.}
    \label{fig:example-2datasets}
\end{figure}

\subsection{Experiments}\label{experiments}

The experiments presented in this work are designed to investigate how the number of MPI affects modeling physical phenomena using GNNs. Our focus is exclusively on assessing the effect of this hyperparameter, which we hypothesize to be critical to the performance and generalization of such architectures. 

To enable a fair comparison across different models, we fix the number of trainable parameters, ensuring that each model maintains sufficient learning capacity for the given tasks. This is achieved through the shared-weight design of the processor module of the model, where the only varying factor among different architectures is the number of MPI. This design choice prevents performance differences from being attributed to the model's expressiveness, thereby isolating the effect of the number of MPI. To validate our hypothesis, we conduct a sensitivity analysis on model performance with respect to the number of MPI. Each configuration is trained using multiple random initialization seeds, allowing us to derive statistically robust insights into the performance trends.

Section~\ref{sec:lower-bound-exp} examines the proposed lower bound for hyperbolic equations, studying the relationship between the distance travelled by the message as it passes through the graph, and the distance physically travelled by the wave whose transmission is modeled by the equation in question. We demonstrate, in the context of the hyperbolic problem, that this relationship can effectively guide the optimal selection of MPI as a hyperparameter in GNNs.

Section~\ref{sec:lower-bound-geom} investigates the proposed lower bound for parabolic and elliptic equations by analyzing the influence of spatial discretization for a domain geometry. In this context, we focus on parabolic and elliptic problems, where the system's behavior is not governed by wavefront propagation, but instead emerges from the global structure of the geometry itself.

%

\subsection{Wave equation}
\label{sec:lower-bound-exp}

For systems governed by hyperbolic equations, as mentioned above, our hypothesis is that the network must send messages that are always ahead of the physical wave modeled by the equation. The sending of messages can never lag behind the physics, or else the affected nodes will not receive the necessary information to be able to reproduce the physical phenomenon, the wave. 

\begin{figure}[h!]
\includegraphics[width=1.0\linewidth]{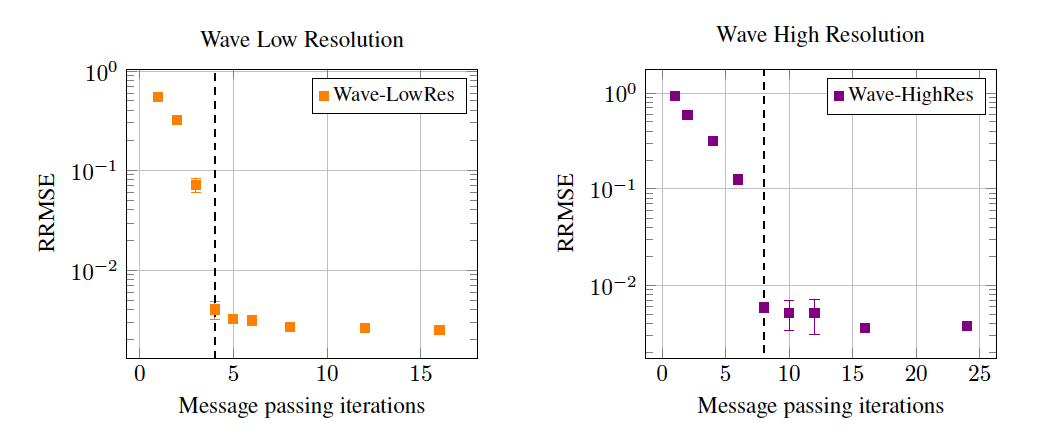}
\caption{
Relative Root Mean Square Error (RRMSE) over three random seeds for \(10\)-step  hyperbolic rollouts, capturing cumulative prediction error. Error bars indicate standard deviation across different model initializations. The dashed vertical lines are the theoretical lower bounds predicted by this work. {Left}: for the low-resolution dataset, the predicted lower bound is 4 MPI.  {Right}: for the high-resolution dataset, the predicted lower bound is 8 message passes per time step. Beyond the threshold, errors converge. All models share the same number of parameters (shared weights), with a fixed latent space size of 256.
}
\label{fig:wave_nstep_error1}
\end{figure}

This relationship establishes a physics-guided lower bound for the hyperparameter settings, ensuring model convergence—defined here as the model's ability to integrate multiple rollout steps without accumulating significant drift or deviating into physically implausible scenarios. This reduces uncertainty during hyperparameter tuning and, consequently, lowers computational cost. 
The experimental results shown in Fig.~\ref{fig:wave_nstep_error1} validate the proposed hypothesis in both scenarios. The plots display the $n$-step rollout error on the test set for varying numbers of MPI across two different lower-bound configurations. As illustrated, when the number of MPI is set below the lower bound (indicated by the dashed line), the models fail to capture the underlying physical behavior, resulting in poor performance. It is only once the lower bound is exceeded (optimal setting) that the models begin to effectively learn the physics, ultimately producing accurate predictions. 

Moreover, an asymptotic behavior can clearly be observed: as the number of MPI enters the optimal regime, the performance tends to saturate, flattening the relationship between prediction error and the number of iterations. This proves that the number of MPI is a crucial hyperparameter in the model's design, and that can be bounded using expert knowledge. Although all models have the same number of trainable parameters, increasing the number of MPI significantly improves problem understanding and convergence—reducing relative errors by up to three orders of magnitude. This highlights that performance gains come not from model size, but from more effective information propagation.

\begin{figure}[h!]
    \centering
    \begin{subfigure}[b]{\linewidth}
        \centering
        \includegraphics[width=1.0\linewidth]{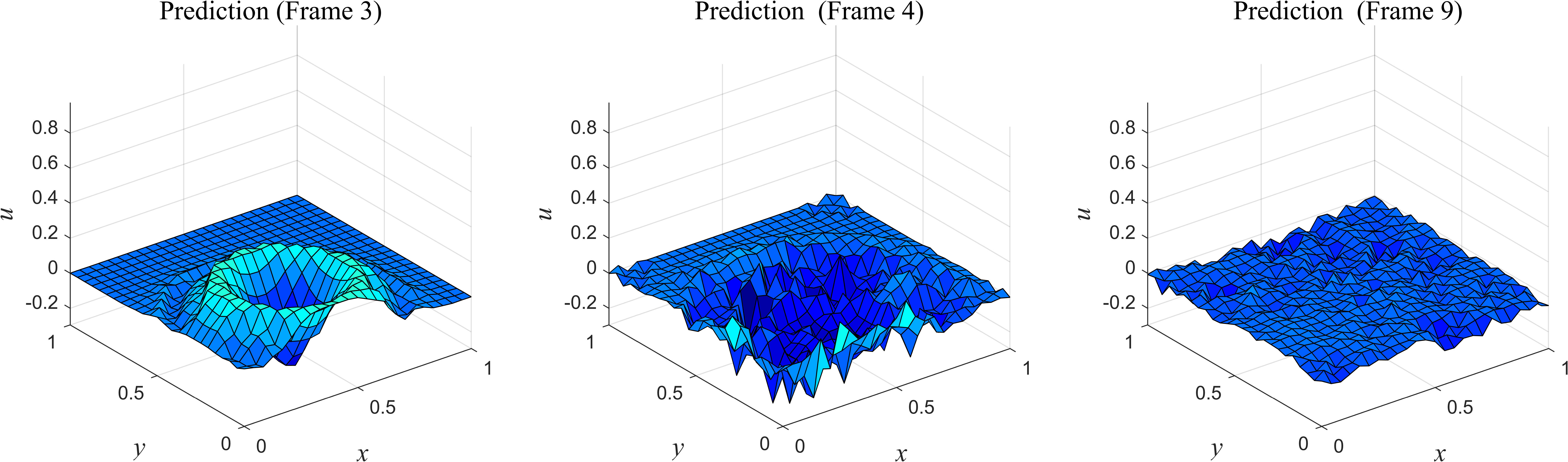}
        \caption{Under-reach condition: number of MPI below the suggested lower bound}
        \label{fig:underreaching}
    \end{subfigure}
    
    \vspace{1em} 
    
    \begin{subfigure}[b]{\linewidth}
        \centering
        \includegraphics[width=1.0\linewidth]{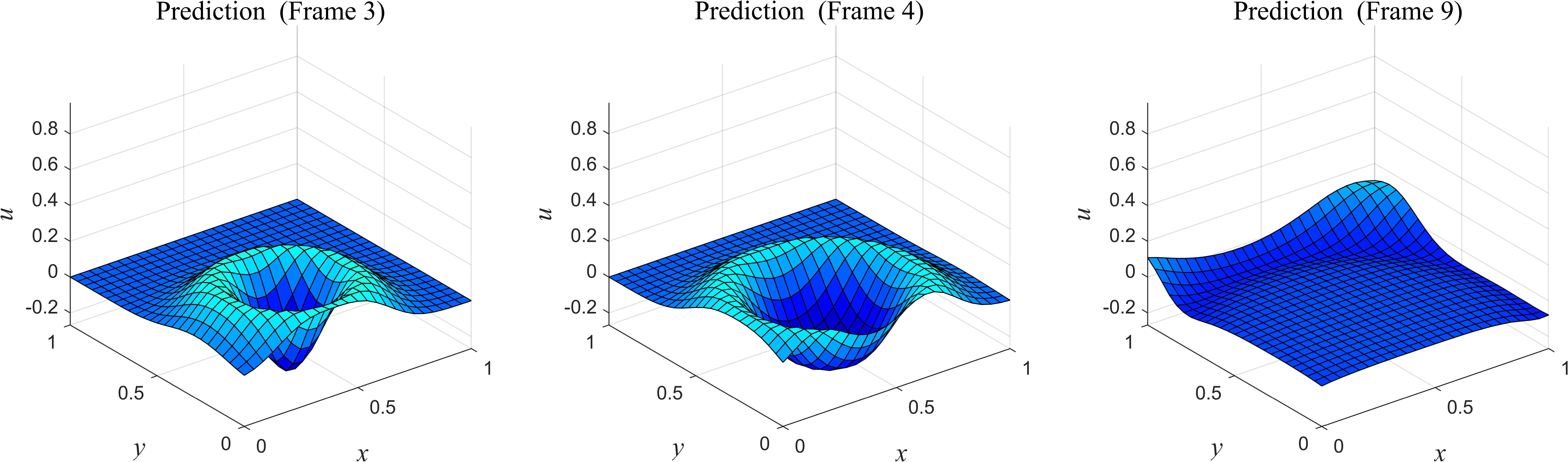}
        \caption{Optimal-reach condition: number of MPI above the suggested lower bound}
        \label{fig:optimalreach}
    \end{subfigure}
    
    \caption{Comparison of GNN prediction performance across different time steps and message passing iterations for the wave problem. Left: early time step ($n=3$), middle: intermediate ($n=4$), right: later stage ($n=9$). (a) With $M=2$ MPI (under-reaching regime, below optimal threshold); (b) $M=8$ MPI (optimal regime, demonstrating prediction accuracy and stable rollout performance).}
    \label{fig:comparison_models_wave}
\end{figure}

The primary reason for poor performance when the model is insufficiently or sub-optimally hyperparameterized is that it is forced to learn the physical phenomenon from incomplete information. This limitation truncates the model's ability to capture the underlying physics, even when the latent space has sufficient expressiveness. As a result, instead of producing meaningful predictions, this can lead to divergence, see Fig. \ref{fig:comparison_models_wave}. 


\subsection{Phenomena governed by parabolic and elliptic PDEs}
\label{sec:lower-bound-geom}

When simulating phenomena governed by parabolic or elliptic PDEs, we hypothesize that message-passing GNN convergence requires each node to eventually acquire a global view of the system state. Since the GNN updates node states solely through the message passing algorithm, as described in Sections~\ref{sec:parabolic_theory}-\ref{sec:elliptic_theory}, this motivates defining the lower bound from a geometric perspective.

\subsubsection{Fourier equation}

\begin{figure}[h!]
\includegraphics[width=1.0\linewidth]{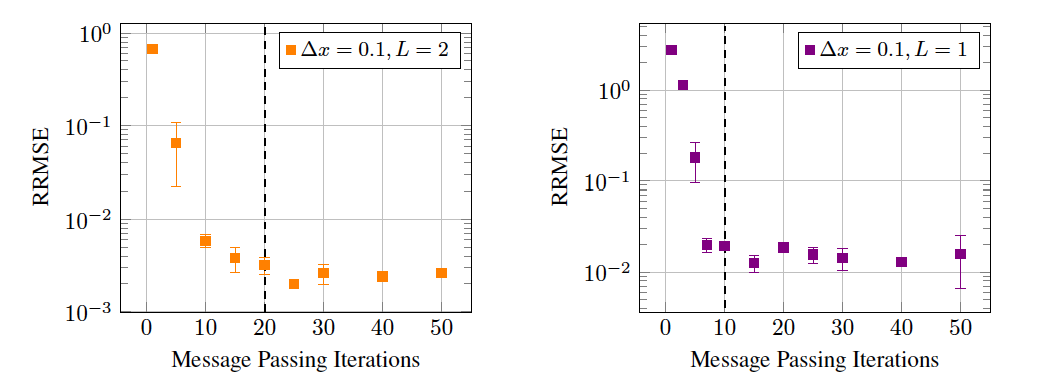}

\caption{
Error over five random seeds for \(n\)-step parabolic rollouts. Dashed lines represent the suggested lower bound. Left: \(2{\times}2\) domain, giving an optimal number of message passing iterations of $M= 20$. Right: \(1{\times}1\) domain, thus giving $M= 10$. Errors remain virtually stable from this point onwards. In any case, for a very large number of passes it seems that the error increases slightly, which may be due to the phenomenon of over-smoothing. All models use a fixed 256-dimensional latent space and share parameters.
}
\label{fig:heat_nstep_error}
\end{figure}

We validate the hypothesis using two different cases, with low and high resolution, respectively. As shown in Fig.~\ref{fig:heat_nstep_error}, an asymptotic behavior is clearly observed: as the number of MPI enters the optimal regime, performance tends to saturate, flattening the relationship between prediction error and the number of iterations. Moreover, increasing the number of iterations beyond this point does not lead to significant reductions in error, indicating saturation and identifying the minimal number of MPI. A high number of MPI brings higher computational cost, which scales as \( \mathcal O(M) \), where \( M \) is the number of MPI.

\begin{figure}[h!]
    \centering
\includegraphics[width=1.0\linewidth]{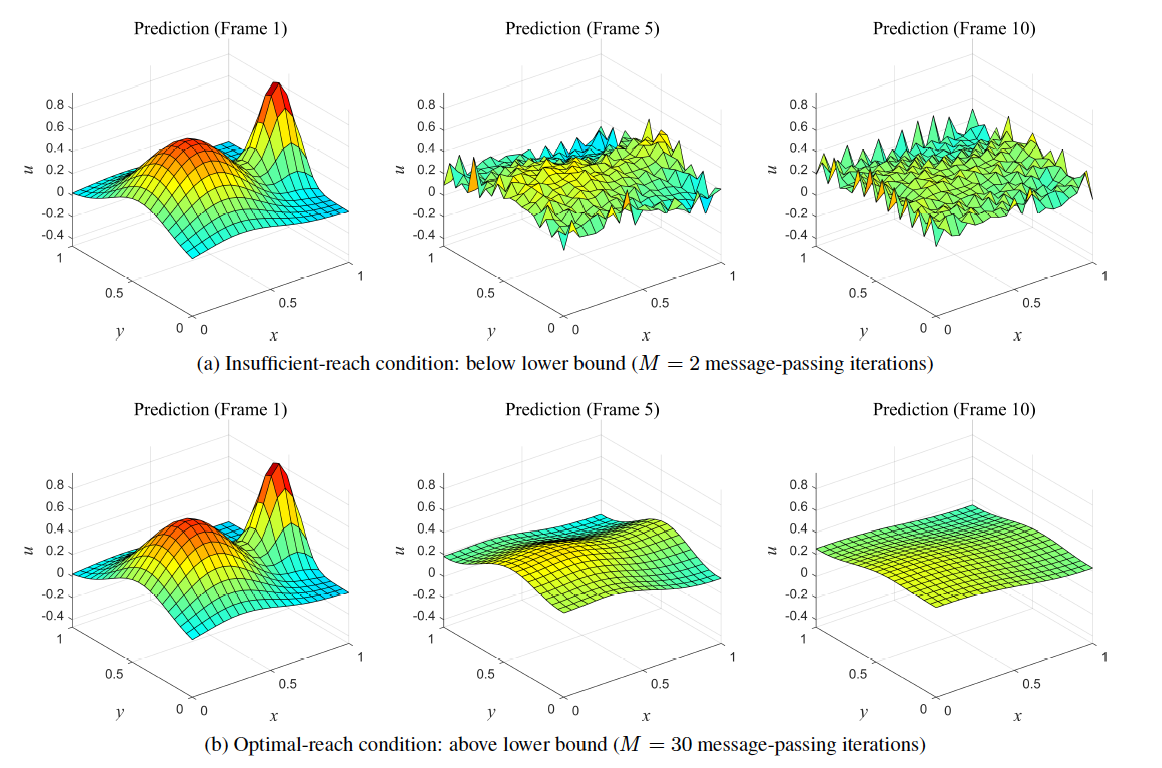}    
\caption{Comparison of GNN performance for the Fourier problem. (a) Optimal design with $M=30$ MPI achieves peak accuracy, demonstrating robust spatial interpretability and stable predictions. (b) Insufficient message passing ($M=2$ MPI) fails to capture the solution dynamics.}
    \label{fig:comparison_models_parabolic}
\end{figure}

Fig.~\ref{fig:comparison_models_parabolic} compares the model's performance between optimal and insufficient message-passing configurations. The results demonstrate that avoiding under-reaching significantly enhances convergence efficiency and mitigates biased or noisy predictions.

\subsubsection{Electrostatic potential problem}

For the electrostatic potential problem, we validate the hypothesis using two different lower-bounded cases, with low- and high-resolution discretisations, that have lower bounds of $M=10$ and $M=20$, respectively. Here, the input to the GNN is a distribution of charge densities $\rho_k$ and the output the resulting potential field $u$. This confirms that the lower-bound behavior is dictated purely by the spatial discretization size and the geometry of the domain. This outcome is expected in elliptic systems, where the solution influences the entire domain instantaneously (i.e., the whole state is effectively updated in a single inference step) and time is not a variable of the problem. 

\begin{figure}[h!]
\includegraphics[width=1.0\linewidth]{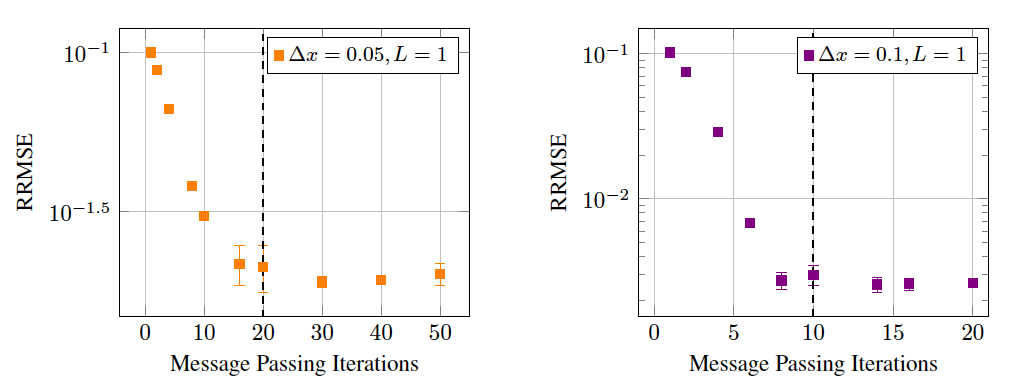}    
\caption{
Relative root mean square error over four random seeds for \(n\)-step parabolic rollouts for the electrostatic potential problem. Dashed lines represent the predicted lower bound. Left: High-resolution dataset with $M= 20$. Right: Low resolution dataset, with $M=10$. Errors converge asymptotically beyond these thresholds. All models use a fixed 256-dimensional latent space and share parameters across number of MPI.
}
\label{fig:poisson_nstep_error}
\end{figure}

\begin{figure}[h!]
    \centering
        \includegraphics[width=0.9\linewidth]{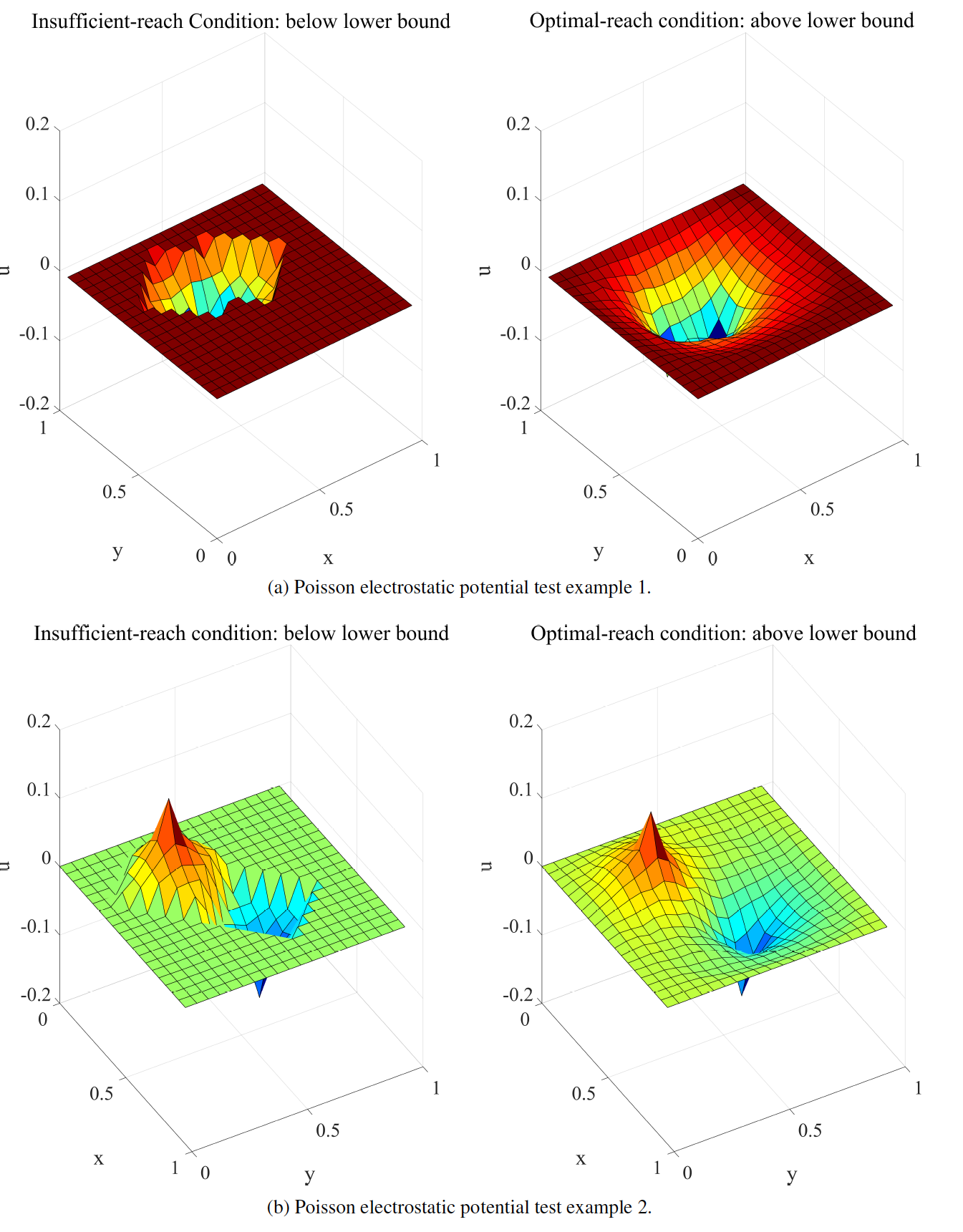}    

    \caption{Comparison of GNN performance as dynamic parabolic integrators, with lower bound set to 20. (a) Insufficient-reach design below the lower bound with only 4 MPI fails. (b) Optimal-reach design above the lower bound with 30 MPI achieves peak accuracy, demonstrating robust spatial interpretability and stable rollout predictions.}
    \label{fig:comparison_models_poisson}
\end{figure}

As shown in Fig.~\ref{fig:poisson_nstep_error}, both datasets exhibit an asymptotic performance trend as the number of MPI increases. Additionally, Fig.~\ref{fig:comparison_models_poisson} illustrates the impact of meeting or violating the geometric lower bound. When the model is under-reached, it fails to capture the global behavior of the physical phenomenon, as its predictions are limited to the region within the restricted message-passing range. In contrast, models configured within the optimal regime effectively propagate information across the entire domain, enabling accurate and physically consistent predictions.

\subsubsection{Incremental forming}

For the incremental forming problem, we employ the MeshGraphNet architecture \cite{pfaff2021learning}. This setup differs from the previously discussed problems in that it involves a multi-graph configuration, where a rigid actuator contacts a deformable plate, and the network is tasked with predicting both displacement and von Mises stress fields. More information about this architecture can be found at \cite{CuetoMikel}.  Despite the increased complexity of the scenario, the core hypothesis regarding MPI remains valid, as the underlying PDE remains elliptic (the problem is assumed quasi-static). 

\begin{figure}[h!]
        \includegraphics[width=0.9\linewidth]{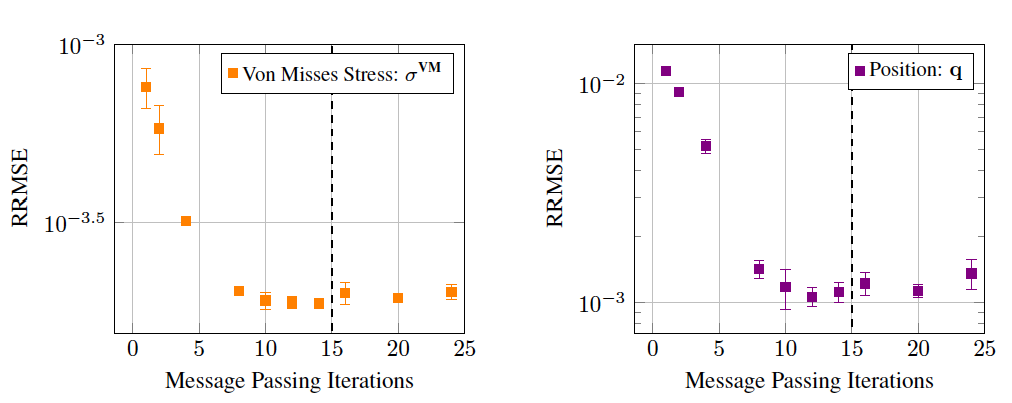}    

\caption{
Error over five random seeds for \(n\)-step predictions in the incremental forming problem, showing cumulative error across rollout. The dashed vertical line represents the suggested lower bound for the number of message passing iterations, $M$. When the model's design is below the lower bound the results reveal higher errors . Both von Mises stress (left) and position (right) errors are shown, each showing a minimum for the predicted optimal MPI value. Models use a 64-dimensional latent space with shared weights across MPI to maintain constant model size.
}

\label{fig:collision_nstep_error}
\end{figure}

As shown in Fig.~\ref{fig:collision_nstep_error}, the multi-graph configuration also exhibits an asymptotic performance trend as the number of MPI approaches the predicted lower bound. It is noted, however, that an increase in the number of passes beyond the predicted figure provides some increase in error, albeit slight. This increase may be due to another phenomenon, known as over-smoothing \cite{qureshi2023limits,jiang2024limiting}. Over-smoothing is a phenomenon whereby, if the number of message passes grows too large, all nodes in the graph end up carrying very similar information, losing their uniqueness and causing a drop in prediction accuracy.

\begin{figure}[h!]
    \centering
            \includegraphics[width=\linewidth]{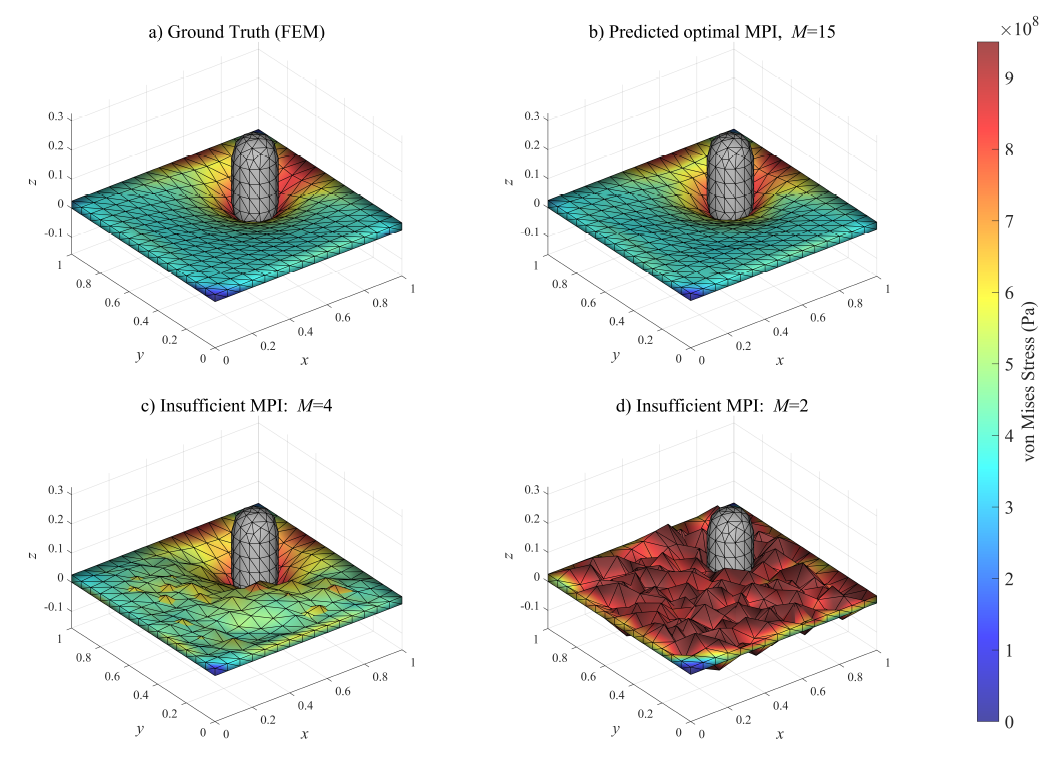}    

    \caption{Comparison of GNN performance as as a function of the number of passes for the incremental forming problem. a) The ground truth FEM solution; b) Prediction produced by the model with 15 message-passing iterations, above the lower bound; c) and d) display the divergence observed when the message-passing iterations does not satisfy the lower bound, with 4 and 2 number of message passing iterations respectively.}
    \label{fig:comparison_models_collision}
\end{figure}

Fig.~\ref{fig:comparison_models_collision} illustrates the effect of correctly configuring the number of MPI. As can be seen, the accuracy of the simulation can be seriously affected if the number of passes is much lower than necessary. As in the preceding cases, setting the number of MPIs to the number predicted by our work ensures excellent accuracy for all the simulations studied.

Another way to analyze and understand the process taking place in the latent space is to represent the value of the latent vector $\bs \xi$ as a function of the number of message passes made. This value provides information that corroborates the analysis carried out so far. The encoded nodal input to the processor has been coined as $\bs{\xi}^h_i \in \mathbb{R}^D$, where $D$ denotes the latent dimension. After performing a message passing iteration, the latent vector is updated to yield the new representation $\bs{\xi}^{h+1}_i$, where $h=1, 2, \ldots, M$ and $i$ represents the node number.  

To normalize this latent vector, we first compute its $L^2$-norm. This scalar field is then divided by the corresponding norm of the encoded input $\bs{\xi}^0_i\in \mathbb{R}$ to express a new relative scalar magnitude,
\begin{equation}\label{norm}
U^m_i =  \frac{\|\bs{\xi}^m_i\|_2}{\|\bs{\xi}^0_i\|_2}.
\end{equation}
The result is a latent scalar value per node in the graph. Therefore, values close to 1 imply that the variation in the latent space for each aggregation is negligible or nearly negligible, indicating no significant relative change with respect to the encoded input. On the other hand, higher values demonstrate greater evolution or change, creating a pseudo-attention map for the graph network.

\begin{figure}[h!]
    \centering
   \includegraphics[width=\linewidth]{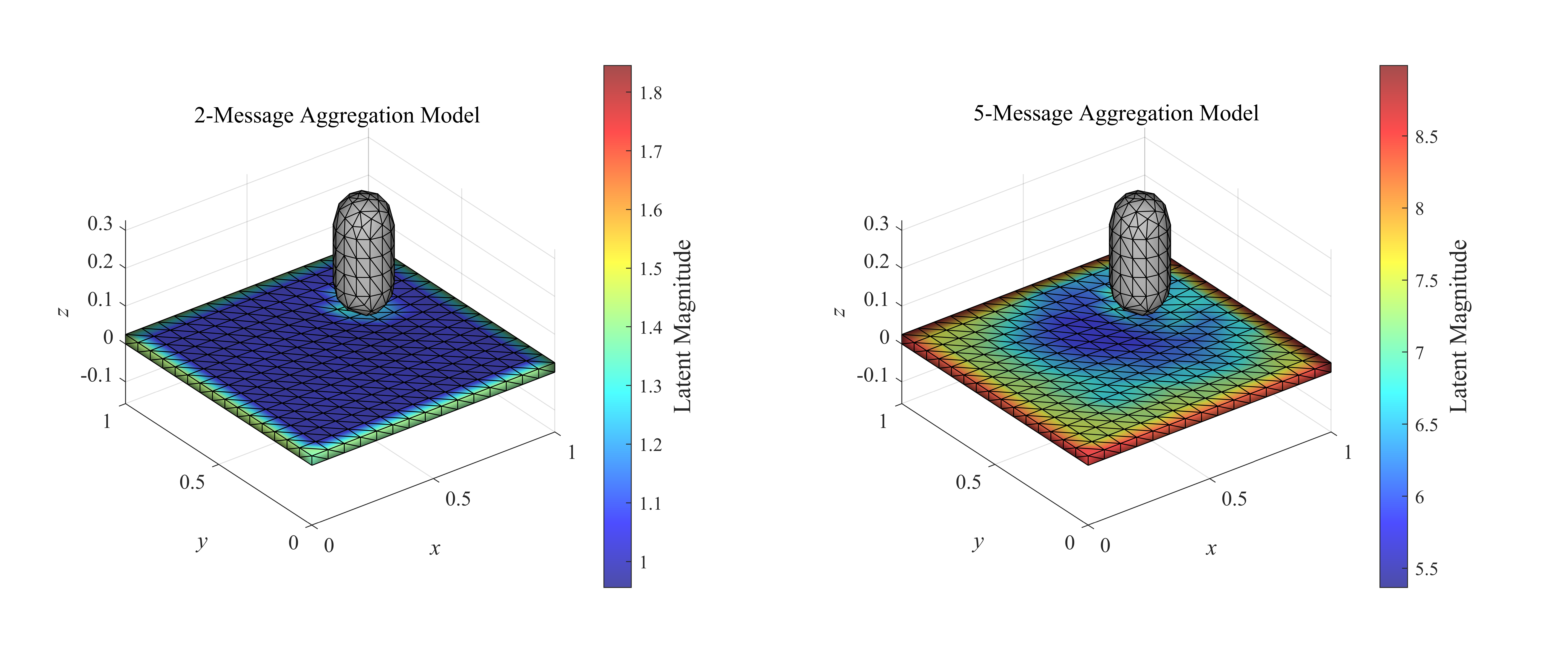};
    \caption{The color map represents the normalized latent vector  magnitude $U^m_i$ at each node $i$, as given by Eq. (\ref{norm}). The analysis compares two configurations: one with 15 message-passing steps (right) and another with only 2 (left), applied during inference on the first input of the incremental forming problem rollout. The resulting latent representation acts as a pseudo-attention map, illustrating how the model distributes focus across the graph structure, mimicking attention-like behavior. Note that the color scales are independently normalized for each plot. }
    \label{fig:heatmap_ellip2}
\end{figure}

For this problem (but similar results can be obtained for the previous problems), the importance of propagating the effect of the tool to the plate boundaries becomes critical. The model with only two message passing steps fails to properly inform central regions. This can alternatively be reflected as low latent magnitudes and limited responsiveness to both the boundaries and the actuator. In contrast, the model with 15 message passing steps produces a fully informed latent representation, with clear relative variation across the entire domain, enabling a more physically accurate and coherent prediction, see Fig. \ref{fig:heatmap_ellip2}.

\subsection{Extrapolation Performance}
\label{sec:extrapolation}

GNN are a powerful tool for handling different geometries as well as varying initial and boundary  conditions during inference, enabling the modeling and extrapolation to unseen scenarios. However, like all deep learning methods, this is usually achieved not without significant effort. It is crucial to understand both the underlying physics of each problem and the technology being used. Only then can we achieve meaningful extrapolation, grounded in the idea that, in many cases, we are essentially performing interpolation within a learned latent space, something at what NN are excellent at.

\begin{figure}[h!]
\centering
            \includegraphics[width=0.6\linewidth]{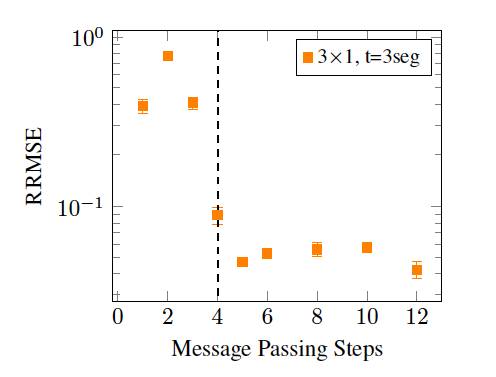}    

\caption{Relative Root Mean Square Error (RRMSE) over a test extrapolation out of the training domain for the wave problem. Error bars indicate standard deviation across different model initializations. Errors are higher in under-reaching regimes—below the message aggregation limit (dotted line: 4 passes for low-res). Beyond this threshold, errors converge, validating the under-reach theory also for extrapolation scenarios. All models share the same number of parameters (shared weights), with a fixed latent space of 64 dimensions.}
\label{fig:EXTRA-WAVES}
\end{figure}

In this work, we use the proposed lower-bound principle to identify scenarios where models are likely to succeed or fail. For hyperbolic problems, geometric variations do not pose a significant challenge due to the local nature of the modeled phenomena---a travelling wave---. As shown in Fig.~\ref{fig:EXTRA-WAVES}, the model generalizes well to unseen geometries and even to longer rollout horizons, demonstrating its ability to extrapolate in both space and time.

\begin{figure}[h!]
    \centering
            \includegraphics[width=\linewidth]{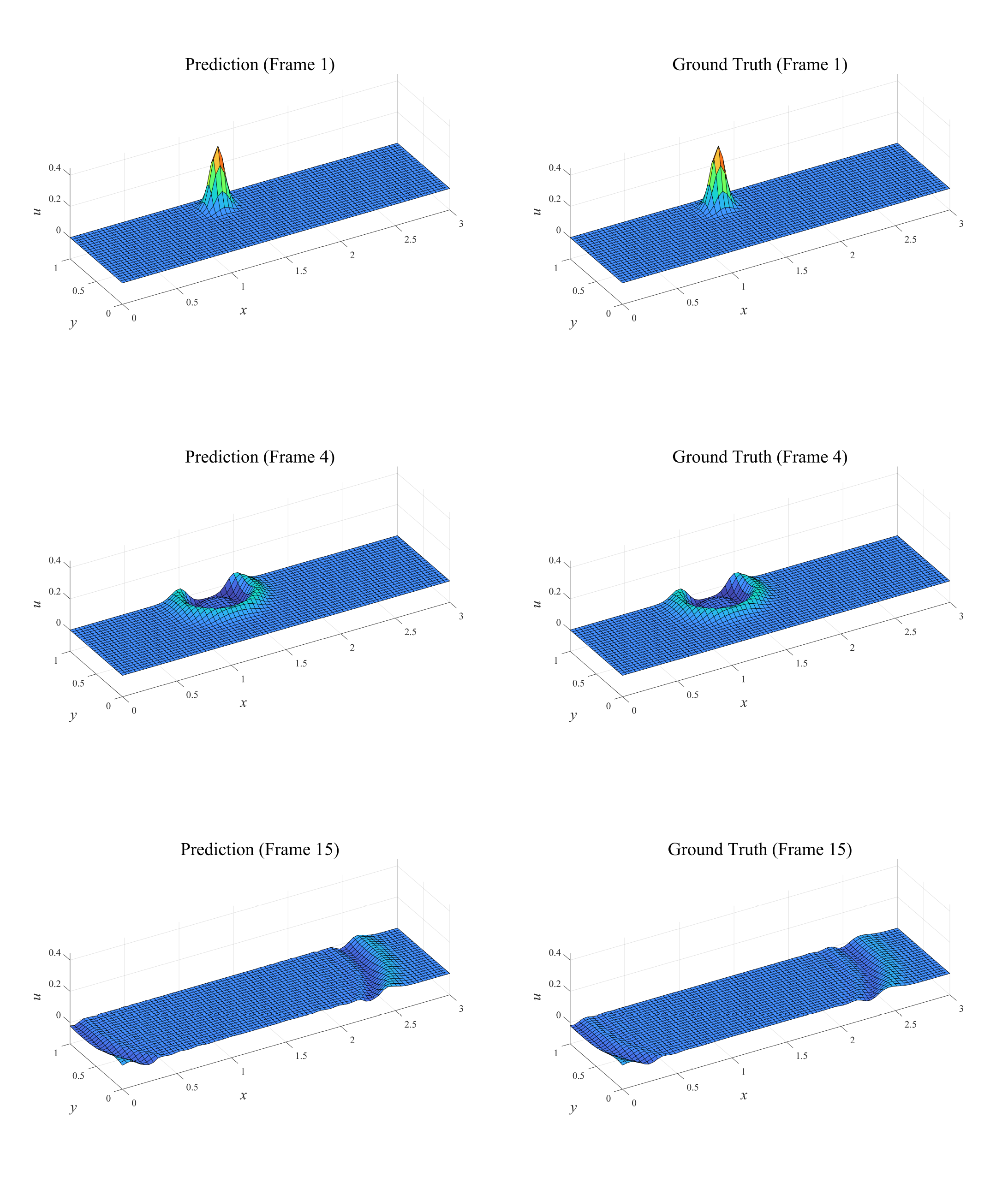}

\caption{A comparison between the dynamic rollout predicted by the Graph Neural Network (GNN) and the numerical solution in an extrapolative inference scenario for the low-resolution wave propagation problem. The model was trained on a $1\times 1$ plate domain and evaluated on an extrapolated $3\times 1$ plate, highlighting its ability in generalizing to larger spatial domains. 5 message passes were employed.}
\label{fig:extrapol}
\end{figure}

\begin{figure}
    \centering
    \includegraphics[trim=0 0 0 50,clip,width=1.0\linewidth]{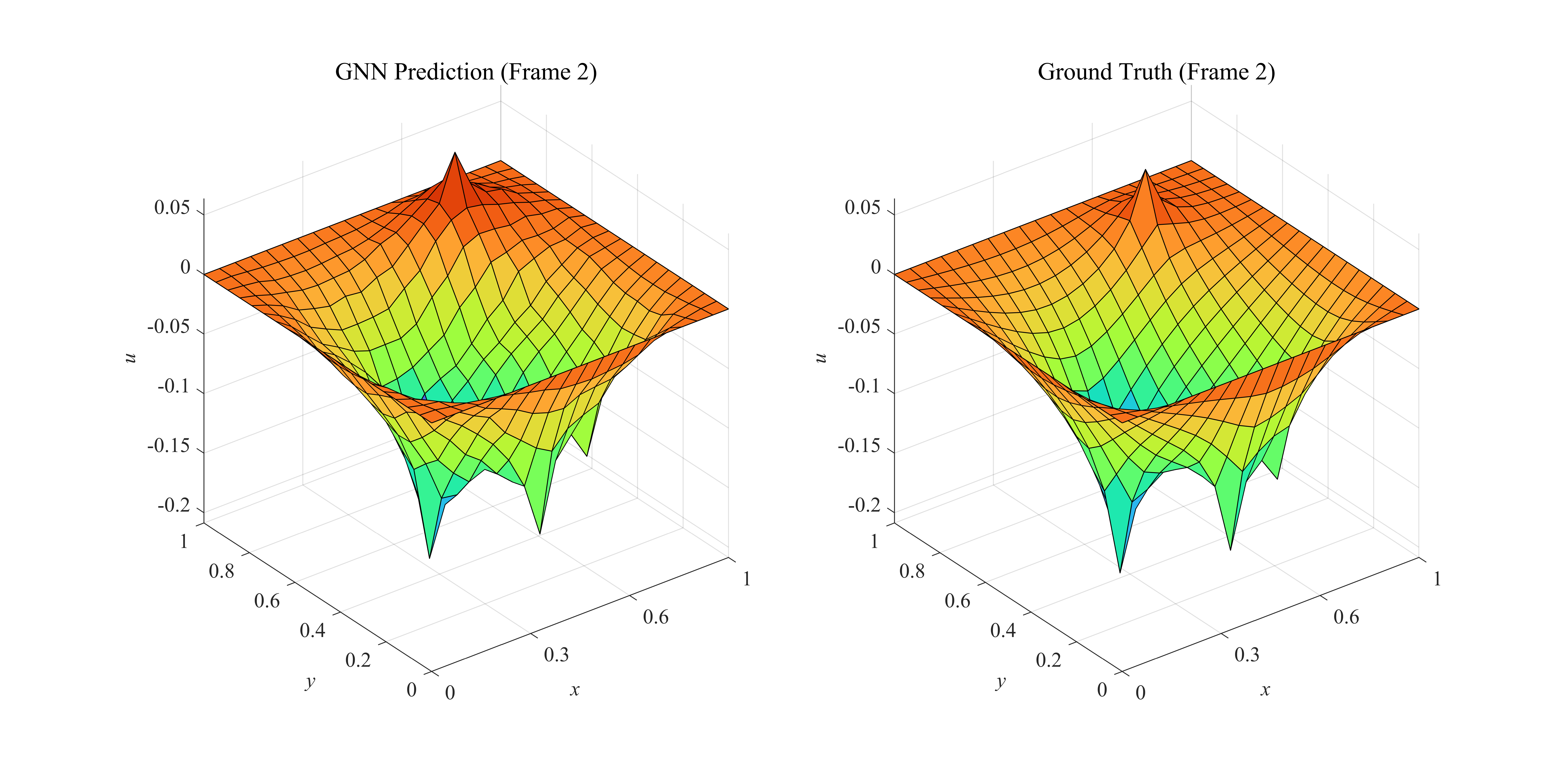}
    \caption{A comparison between the dynamic rollout predicted by the Graph Neural Network (GNN) and the numerical solution in an extrapolative inference setting for the high-resolution Poisson problem. In this case, the trained domain was extended from three initial source points to five, demonstrating the model’s capacity to generalize beyond its training configuration.}
    \label{fig:extrapol2}
\end{figure}

Conversely, for elliptic and parabolic problems, successful extrapolation to unseen scenarios depends on whether the MPI span the entire domain. Fig.~\ref{fig:extrapol} shows that a model trained on the Poisson high-resolution mesh  dataset with a maximum of 3 charges can successfully solve an unseen case with 5 charges. However, when the domain itself changes, from $1\times 1$ to $3\times 3$ as in Fig.~\ref{fig:extrapol2}, the GNN fails to cover the entire domain. Despite the simpler load configuration (only 3 charges), the model's MPI design no longer satisfies the new lower-bound condition and becomes insufficient.

\section{Conclusions}\label{conclusions}

This study systematically investigates the influence of message-passing iterations in graph neural networks when modeling physical systems governed by partial differential equations. Our findings demonstrate that message passing is not merely a computational mechanism but a physics-aligned process that directly impacts model accuracy, generalization, and stability. Specifically, we identify and validate physics-guided lower bounds for message propagation that can be used to inform optimal model design.

For hyperbolic systems, we propose a lower bound based on the relationship between the distance travelled by the message sent and by the physical wave whose evolution is to be resolved. Empirical results from wave propagation datasets confirm that convergence is only achieved when message passing travels ahead of the physical wave. This validates message passing as a critical hyperparameter and establishes a novel, principled approach for hyperparameter tuning in GNNs.

For parabolic and elliptic systems, we demonstrate that convergence requires global information propagation across the domain. This leads us to define a lower bound on message-passing steps, dependent on mesh topology and the geometry of the domain. Across heat diffusion and electrostatic potential datasets, we observe that performance saturates once this bound is reached. This behavior persists even in complex scenarios such as plastic deformation in incremental forming. Notably, even under varying geometries and boundary conditions, the optimal number of iterations remains tightly linked to the spatial discretization, confirming the universality of the proposed criterion.

These experiments demonstrate that the extrapolation capability of GNNs strongly depends on the type of lower-bound condition governing the problem. For hyperbolic problems, the locality of the solution allows the model to generalize well to unseen geometric domains. For elliptic and parabolic problems, success requires that MPIs span the entire domain at each time step. When this is not the case---as in larger, unseen domains---the model fails due to limited information propagation. However, it performs well on unseen scenarios as long as the domain remains within the reach of the MPIs. 


Overall, our work offers a rigorous framework for physics-aware GNN design, where the number of message-passing iterations is explicitly tied to the physical and geometric properties of the modeled system. These findings provide guidelines for architecture selection and hyperparameter tuning, advancing the use of GNNs in scientific machine learning and data-driven physics simulation.

\section*{Acknowledgements}

This work was supported by the Spanish Ministry of Science and Innovation, AEI/10.13039/501100011033, through Grant number PID2023-147373OB-I00, and by the Ministry for Digital Transformation and the Civil Service, through the ENIA 2022 Chairs for the creation of university-industry chairs in AI, through Grant TSI-100930-2023-1.

This material is also based upon work supported in part by the Army Research Laboratory and the Army Research Office under contract/grant number W911NF2210271.

The authors also acknowledge the support of ESI Group through the chair at the
University of Zaragoza.

\bibliographystyle{unsrt}
\bibliography{references} 

\begin{thebibliography}{10}

\bibitem{montans2023machine}
Francisco~J Mont{\'a}ns, El{\'\i}as Cueto, and Klaus-J{\"u}rgen Bathe.
\newblock Machine learning in computer aided engineering.
\newblock In {\em Machine Learning in Modeling and Simulation: Methods and
  Applications}, pages 1--83. Springer, 2023.

\bibitem{pfaff2021learning}
Tobias Pfaff, Meire Fortunato, Alvaro Sanchez-Gonzalez, and Peter Battaglia.
\newblock Learning mesh-based simulation with graph networks.
\newblock In {\em International Conference on Learning Representations}, 2021.

\bibitem{aldakheel2025physics}
Fadi Aldakheel, Elsayed~S Elsayed, Yousef Heider, and Oliver Weeger.
\newblock Physics-based machine learning for computational fracture mechanics.
\newblock {\em Machine Learning for Computational Science and Engineering},
  1(1):18, 2025.

\bibitem{FlowGNN}
Rishov Sarkar, Stefan Abi-Karam, Yuqi He, Lakshmi Sathidevi, and Cong Hao.
\newblock Flowgnn: A dataflow architecture for real-time workload-agnostic
  graph neural network inference.
\newblock In {\em 2023 IEEE International Symposium on High-Performance
  Computer Architecture (HPCA)}, pages 1099--1112, 2023.

\bibitem{InverseGal}
Han Gao, Matthew~J. Zahr, and Jian-Xun Wang.
\newblock Physics-informed graph neural galerkin networks: A unified framework
  for solving pde-governed forward and inverse problems.
\newblock {\em Computer Methods in Applied Mechanics and Engineering},
  390:114502, 2022.

\bibitem{lam2022graphcast}
R.~Lam, A.~Sanchez-Gonzalez, M.~Willson, P.~Wirnsberger, M.~Fortunato, F.~Alet,
  and P.~Battaglia.
\newblock Graphcast: Learning skillful medium-range global weather forecasting.
\newblock {\em arXiv preprint arXiv:2212.12794}, 2022.

\bibitem{jordan1965calculus}
Károly Jordán.
\newblock {\em Calculus of Finite Differences}, volume~33.
\newblock American Mathematical Society, 1965.

\bibitem{bathe2007finite}
Klaus~Jurgen Bathe.
\newblock Finite element method.
\newblock {\em Wiley encyclopedia of computer science and engineering}, pages
  1--12, 2007.

\bibitem{eymard2000finite}
Robert Eymard, Thierry Gallou{\"e}t, and Rapha{\`e}le Herbin.
\newblock Finite volume methods.
\newblock {\em Handbook of numerical analysis}, 7:713--1018, 2000.

\bibitem{Karniadakis2021}
George~Em Karniadakis, Ioannis~G. Kevrekidis, Lu~Lu, Paris Perdikaris, Sifan
  Wang, and Liu Yang.
\newblock Physics-informed machine learning.
\newblock {\em Nature Reviews Physics}, 3(6):422--440, 2021.

\bibitem{Zhao2024}
Yingxue Zhao, Haoran Li, Haosu Zhou, Hamid~Reza Attar, Tobias Pfaff, and Nan
  Li.
\newblock A review of graph neural network applications in mechanics-related
  domains.
\newblock {\em Artificial Intelligence Review}, 57(11):315, 10 2024.

\bibitem{allen2022physicaldesignusingdifferentiable}
Kelsey~R. Allen, Tatiana Lopez-Guevara, Kimberly Stachenfeld, Alvaro
  Sanchez-Gonzalez, Peter Battaglia, Jessica Hamrick, and Tobias Pfaff.
\newblock Physical design using differentiable learned simulators, 2022.

\bibitem{allen2022learningrigiddynamicsface}
Kelsey~R. Allen, Yulia Rubanova, Tatiana Lopez-Guevara, William Whitney, Alvaro
  Sanchez-Gonzalez, Peter Battaglia, and Tobias Pfaff.
\newblock Learning rigid dynamics with face interaction graph networks, 2022.

\bibitem{CuetoMikel}
Alicia Tierz, Mikel~M. Iparraguirre, Icíar Alfaro, David González, Francisco
  Chinesta, and Elías Cueto.
\newblock On the feasibility of foundational models for the simulation of
  physical phenomena.
\newblock {\em International Journal for Numerical Methods in Engineering},
  126(6):e70027, 2025.

\bibitem{Cuomo2022}
Salvatore Cuomo, Vincenzo~Schiano Di~Cola, Fabio Giampaolo, Gianluigi Rozza,
  Maziar Raissi, and Francesco Piccialli.
\newblock Scientific machine learning through physics--informed neural
  networks: Where we are and what’s next.
\newblock {\em Journal of Scientific Computing}, 92(3):88, 2022.

\bibitem{Cai2021}
Shengze Cai, Zhiping Mao, Zhicheng Wang, Minglang Yin, and George~Em
  Karniadakis.
\newblock Physics-informed neural networks (pinns) for fluid mechanics: a
  review.
\newblock {\em Acta Mechanica Sinica}, 37(12):1727--1738, 2021.

\bibitem{Kim_Lee_Lee_Jhin_Park_2021}
Jungeun Kim, Kookjin Lee, Dongeun Lee, Sheo~Yon Jhin, and Noseong Park.
\newblock Dpm: A novel training method for physics-informed neural networks in
  extrapolation.
\newblock 35:8146--8154, May 2021.

\bibitem{fesser2023understandingmitigatingextrapolationfailures}
Lukas Fesser, Luca D'Amico-Wong, and Richard Qiu.
\newblock Understanding and mitigating extrapolation failures in
  physics-informed neural networks, 2023.

\bibitem{Fresca2021}
Stefania Fresca, Luca Dede’, and Andrea Manzoni.
\newblock A comprehensive deep learning-based approach to reduced order
  modeling of nonlinear time-dependent parametrized pdes.
\newblock {\em Journal of Scientific Computing}, 87(2):61, 2021.

\bibitem{fresca-PODL}
Stefania Fresca and Andrea Manzoni.
\newblock Pod-dl-rom: Enhancing deep learning-based reduced order models for
  nonlinear parametrized pdes by proper orthogonal decomposition.
\newblock {\em Computer Methods in Applied Mechanics and Engineering},
  388:114181, 2022.

\bibitem{Fresca2025}
Simone Brivio, Stefania Fresca, and Andrea Manzoni.
\newblock Handling geometrical variability in nonlinear reduced order modeling
  through continuous geometry-aware dl-roms.
\newblock {\em Computer Methods in Applied Mechanics and Engineering},
  442:117989, 2025.

\bibitem{lecun1998gradient}
Yann LeCun, L{\'e}on Bottou, Yoshua Bengio, and Patrick Haffner.
\newblock Gradient-based learning applied to document recognition.
\newblock {\em Proceedings of the {IEEE}}, 86(11):2278--2324, 1998.

\bibitem{Bronstein-2021_geodistnce}
Michael~M. Bronstein, Joan Bruna, Taco Cohen, and Petar Veličković.
\newblock {\em Geometric Deep Learning: Grids, Groups, Graphs, Geodesics, and
  Gauges}, chapter GeometricDomains: the 5Gs, page~50.
\newblock arXiv, 2021.

\bibitem{bronstein2017geometric}
Michael~M. Bronstein, Joan Bruna, Yann LeCun, Arthur Szlam, and Pierre
  Vandergheynst.
\newblock Geometric deep learning: Going beyond euclidean data.
\newblock {\em {IEEE Signal Processing Magazine}}, 34(4):18--42, 2017.

\bibitem{GNNsurvey}
Zonghan Wu, Shirui Pan, Fengwen Chen, Guodong Long, Chengqi Zhang, and
  Philip~S. Yu.
\newblock A comprehensive survey on graph neural networks.
\newblock {\em IEEE Transactions on Neural Networks and Learning Systems},
  32(1):4--24, 2021.

\bibitem{NEURIPS2022_17b598fd}
Abishek Thangamuthu, Gunjan Kumar, Suresh Bishnoi, Ravinder Bhattoo, N~M~Anoop
  Krishnan, and Sayan Ranu.
\newblock Unravelling the performance of physics-informed graph neural networks
  for dynamical systems.
\newblock In S.~Koyejo, S.~Mohamed, A.~Agarwal, D.~Belgrave, K.~Cho, and A.~Oh,
  editors, {\em Advances in Neural Information Processing Systems}, volume~35,
  pages 3691--3702. Curran Associates, Inc., 2022.

\bibitem{LIU2023211486}
Wendi Liu and Michael~J. Pyrcz.
\newblock Physics-informed graph neural network for spatial-temporal production
  forecasting.
\newblock {\em Geoenergy Science and Engineering}, 223:211486, 2023.

\bibitem{TIGNNs}
Quercus Hernandez, Alberto Badias, Francisco Chinesta, and Elias Cueto.
\newblock Thermodynamics-informed graph neural networks.
\newblock {\em IEEE Transactions on Artificial Intelligence}, 2022.

\bibitem{hernandez2021structure}
Quercus Hern{\'a}ndez, Alberto Bad{\'\i}as, David Gonz{\'a}lez, Francisco
  Chinesta, and El{\'\i}as Cueto.
\newblock Structure-preserving neural networks.
\newblock {\em Journal of Computational Physics}, 426:109950, 2021.

\bibitem{hernandez2023port}
Quercus Hern{\'a}ndez, Alberto Bad{\'\i}as, Francisco Chinesta, and El{\'\i}as
  Cueto.
\newblock Port-metriplectic neural networks: thermodynamics-informed machine
  learning of complex physical systems.
\newblock {\em Computational Mechanics}, 72(3):553--561, 2023.

\bibitem{hernandez2023thermodynamics}
Quercus Hern{\'a}ndez, Alberto Bad{\'\i}as, Francisco Chinesta, and El{\'\i}as
  Cueto.
\newblock Thermodynamics-informed neural networks for physically realistic
  mixed reality.
\newblock {\em Computer Methods in Applied Mechanics and Engineering},
  407:115912, 2023.

\bibitem{tierz2025graph}
Alicia Tierz, Iciar Alfaro, David Gonz{\'a}lez, Francisco Chinesta, and
  El{\'\i}as Cueto.
\newblock Graph neural networks informed locally by thermodynamics.
\newblock {\em Engineering Applications of Artificial Intelligence},
  144:110108, 2025.

\bibitem{gruber2023reversible}
Anthony Gruber, Kookjin Lee, and Nathaniel Trask.
\newblock Reversible and irreversible bracket-based dynamics for deep graph
  neural networks.
\newblock {\em Advances in Neural Information Processing Systems},
  36:38454--38484, 2023.

\bibitem{moya2023thermodynamics}
Beatriz Moya, Alberto Bad{\'\i}as, David Gonz{\'a}lez, Francisco Chinesta, and
  El{\'\i}as Cueto.
\newblock A thermodynamics-informed active learning approach to perception and
  reasoning about fluids.
\newblock {\em Computational Mechanics}, 72(3):577--591, 2023.

\bibitem{bermejo2024thermodynamics}
Carlos Bermejo-Barbanoj, Beatriz Moya, Alberto Bad{\'\i}as, Francisco Chinesta,
  and El{\'\i}as Cueto.
\newblock Thermodynamics-informed super-resolution of scarce temporal dynamics
  data.
\newblock {\em Computer Methods in Applied Mechanics and Engineering},
  430:117210, 2024.

\bibitem{FeurerHutter}
Matthias Feurer and Frank Hutter.
\newblock {\em Hyperparameter Optimization}, pages 3--33.
\newblock 05 2019.

\bibitem{pmlr-v139-balcilar21a}
Muhammet Balcilar, Pierre Heroux, Benoit Gauzere, Pascal Vasseur, Sebastien
  Adam, and Paul Honeine.
\newblock Breaking the limits of message passing graph neural networks.
\newblock In Marina Meila and Tong Zhang, editors, {\em Proceedings of the 38th
  International Conference on Machine Learning}, volume 139 of {\em Proceedings
  of Machine Learning Research}, pages 599--608. PMLR, 18--24 Jul 2021.

\bibitem{lax1973hyperbolic}
Peter~D. Lax.
\newblock {\em Hyperbolic Systems of Conservation Laws and the Mathematical
  Theory of Shock Waves}.
\newblock Number~11 in Conference Board of the Mathematical Sciences Regional
  Conference Series in Applied Mathematics. Society for Industrial and Applied
  Mathematics, Philadelphia, PA, 1973.

\bibitem{auriault2017paradox}
J-L Auriault.
\newblock The paradox of fourier heat equation: A theoretical refutation.
\newblock {\em International Journal of Engineering Science}, 118:82--88, 2017.

\bibitem{Gilmer2020}
Justin Gilmer, Samuel~S. Schoenholz, Patrick~F. Riley, Oriol Vinyals, and
  George~E. Dahl.
\newblock {\em Message Passing Neural Networks}, pages 199--214.
\newblock Springer International Publishing, Cham, 2020.

\bibitem{lu2024nodemixup}
Weigang Lu, Ziyu Guan, Wei Zhao, Yaming Yang, and Long Jin.
\newblock Nodemixup: Tackling under-reaching for graph neural networks.
\newblock In {\em Proceedings of the AAAI Conference on Artificial
  Intelligence}, volume~38, pages 14175--14183, 2024.

\bibitem{brandstetter2022message}
Johannes Brandstetter, Daniel Worrall, and Max Welling.
\newblock Message passing neural pde solvers.
\newblock {\em arXiv preprint arXiv:2202.03376}, 2022.

\bibitem{de2013courant}
Carlos~A De~Moura and Carlos~S Kubrusly.
\newblock {The Courant--Friedrichs--Lewy (CFL) condition}.
\newblock {\em AMC}, 10(12):45--90, 2013.

\bibitem{bertola2007speed}
V~Bertola and E~Cafaro.
\newblock On the speed of heat.
\newblock {\em Physics Letters A}, 372(1):1--4, 2007.

\bibitem{toppingunderstanding}
Jake Topping, Francesco Di~Giovanni, Benjamin~Paul Chamberlain, Xiaowen Dong,
  and Michael~M Bronstein.
\newblock Understanding over-squashing and bottlenecks on graphs via curvature.
\newblock In {\em International Conference on Learning Representations}, 2021.

\bibitem{qureshi2023limits}
Shaima Qureshi et~al.
\newblock Limits of depth: Over-smoothing and over-squashing in gnns.
\newblock {\em Big Data Mining and Analytics}, 7(1):205--216, 2023.

\bibitem{jiang2024limiting}
Yuanhong Jiang, Dongmian Zou, Xiaoqun Zhang, and Yu~Guang Wang.
\newblock Limiting over-smoothing and over-squashing of graph message passing
  by deep scattering transforms.
\newblock {\em arXiv preprint arXiv:2407.06988}, 2024.

\end{thebibliography}

\end{document}